\documentclass[runningheads]{llncs}

\usepackage{eccv}
\usepackage{eccvabbrv}

\usepackage{graphicx}
\usepackage{booktabs}
\usepackage{multirow}
\usepackage{array}
\usepackage{wrapfig}
\usepackage{lipsum}
\usepackage{arydshln}
\usepackage{nicematrix}
\usepackage{tikz}
\usepackage{orcidlink}

\usepackage{makecell}
\newcolumntype{C}[1]{>{\centering\let\newline\\\arraybackslash\hspace{0pt}}m{#1}}

\usepackage[accsupp]{axessibility}

\usepackage{hyperref}

\def\our{SMAL-pets}

\begin{document}

\title{\our{}: SMAL Based Avatars of Pets from Single Image} 

\titlerunning{SMAL-pets}

\author{Piotr Borycki*\inst{1} \and
Joanna Waczyńska*\inst{1} \and
Yizhe Zhu\inst{2} \and
Yongqiang Gao\inst{2} \and
Przemysław Spurek \inst{1}
}

\institute{Jagiellonian University \and Huawei\\
*Authors contributed to the manuscript equally}

\authorrunning{P. Borycki et al.}

\maketitle

\begin{figure}
\vspace{-0.9cm}
\begin{center}
\includegraphics[width=\linewidth,  trim={0, 80, 20, 0}, clip]{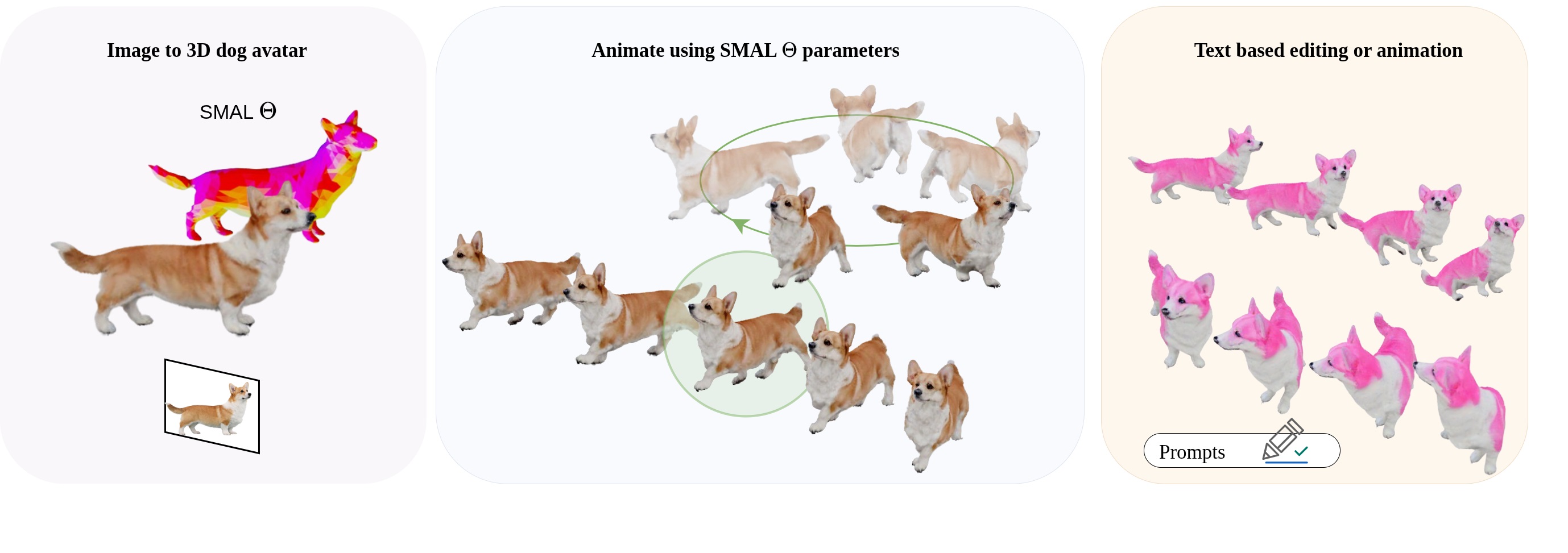}
\end{center}
\vspace{-0.25cm}
    \caption{ We present a novel framework \our{} for generating a fully animatable dog avatar from a single image. Our method achieves state-of-the-art performance by integrating a parametric SMAL mesh with Gaussians primitives. The parametric mesh provides explicit structural control, enabling pose manipulation through interpretable position parameters. We show that our approach supports text-driven avatar animation, enabling users to modify motion and appearance via natural-language prompts.}
    \vspace{-1.3cm}
    \label{fig:teaser}
\end{figure}

\begin{abstract}

Creating high-fidelity, animatable 3D dog avatars remains a formidable challenge in computer vision. Unlike human digital doubles, animal reconstruction faces a critical shortage of large-scale, annotated datasets for specialized applications. Furthermore, the immense morphological diversity across species, breeds, and crosses, which varies significantly in size, proportions, and features, complicates the generalization of existing models. Current reconstruction methods often struggle to capture realistic fur textures. Additionally, ensuring these avatars are fully editable and capable of performing complex, naturalistic movements typically necessitates labor-intensive manual mesh manipulation and expert rigging.
This paper introduces \textbf{\our{}}, a comprehensive framework that generates high-quality, editable animal avatars from a single input image. Our approach bridges the gap between reconstruction and generative modeling by leveraging a hybrid architecture. Our method integrates 3D Gaussian Splatting with the SMAL parametric model to provide a representation that is both visually high-fidelity and anatomically grounded. We introduce a multimodal editing suite that enables users to refine the avatar's appearance and execute complex animations through direct textual prompts. By allowing users to control both the aesthetic and behavioral aspects of the model via natural language, \our{} provides a flexible, robust tool for animation and virtual reality.  

\end{abstract}

\section{Introduction}

With the rapid development of remote collaboration platforms and immersive virtual environments, avatars have evolved from decorative substitutes to essential carriers of digital identity. In modern video games, avatars have become indispensable as they represent the player in the virtual world. Previous research on interaction and gaming has shown that avatars increase player engagement, agency, and emotional connection~\cite{8263407, TENG2017601}.

For human avatars, parametric body mesh models such as SMPL (Skinned Multi-Person Linear model)\cite{SMPl, SMPL-X:2019} have become the industry standard for representing articulated human 3D geometry and motion. In contrast, the creation of high-quality, animatable 3D animal avatars is still a longstanding goal in computer vision with applications ranging from digital film production to virtual reality. Models such as SMAL (Skinned Multi-Animal Linear)~\cite{zuffi20173d}, offer a solution by providing an anatomically grounded parametric mesh. 

Despite the rapid development in the field of human avatars, animal reconstruction faces challenges~\cite{wu2023magicpony}. Creating a realistic aniaml avatar is technically challenging due to visual, behavioral, and performance constraints, where available data is severely limited~\cite{sun2024ponymation}.  Most often, datasets contain dynamic mono camera videos. This makes it impossible to capture the dog from every angle and in every pose, moreover, often they have limited quality (e.g., blurry, low resolution)~\cite{lei2024gart, sinha2023cop3d, wang2025dogmo}. As a result, the methods often require the generation of additional views. Existing works in the literature point out that limited quality in animal subjects~\cite{zhou2025stable}. Large generative methods can generate 3D models, but it is observed that the obtained appearance is often artificial, cartoon or plastic-like, and artifacts may appear~\cite{lei2024gart, li2026animalgs, BORJI2023104771, niewiadomski2025ICCV}. 

Accurately fitting SMAL to a moving animal remains a challenging task~\cite{niewiadomski2025ICCV, wang2025dogmo, lei2024gart, ruegg2023bite}. The limited number of mesh faces restricts the model’s ability to capture fine-grained geometric details and realistic appearance. A significant shift has occurred with the rise of 3D Gaussian Splatting (3DGS)~\cite{kerbl3Dgaussians}. Techniques like GART~\cite{lei2024gart} and AnimalGS~\cite{li2026animalgs} leverage 3DGS for its superior rendering capabilities. However, the generated avatar and the editing method are still difficult to control. To address this, recent research has explored integrating GS with parametric priors, such as GaussianMorphing~\cite{li2025gaussianmorphing}, or using dynamic meshes, as seen in ActionMesh~\cite{sabathier2026actionmesh} and Artemis~\cite{luo2022artemis}.

In this paper, we introduce \textbf{\our{}}, a comprehensive framework designed to solve these limitations by providing an end-to-end tool for creating animatable animal avatars from a single image. Our method acts as a hybrid bridge between generative modeling and parametric reconstruction. By integrating 3D Gaussian Splatting with the SMAL model, we ensure that the resulting avatar is both visually pleasing and structurally sound. \our{} leverages Gaussian Splatting representation to enable semantic editing of textures and shape via natural-language prompts, see Fig.~\ref{fig:teaser}.

\begin{figure}[t]
    \centering
    \includegraphics[width=\linewidth]{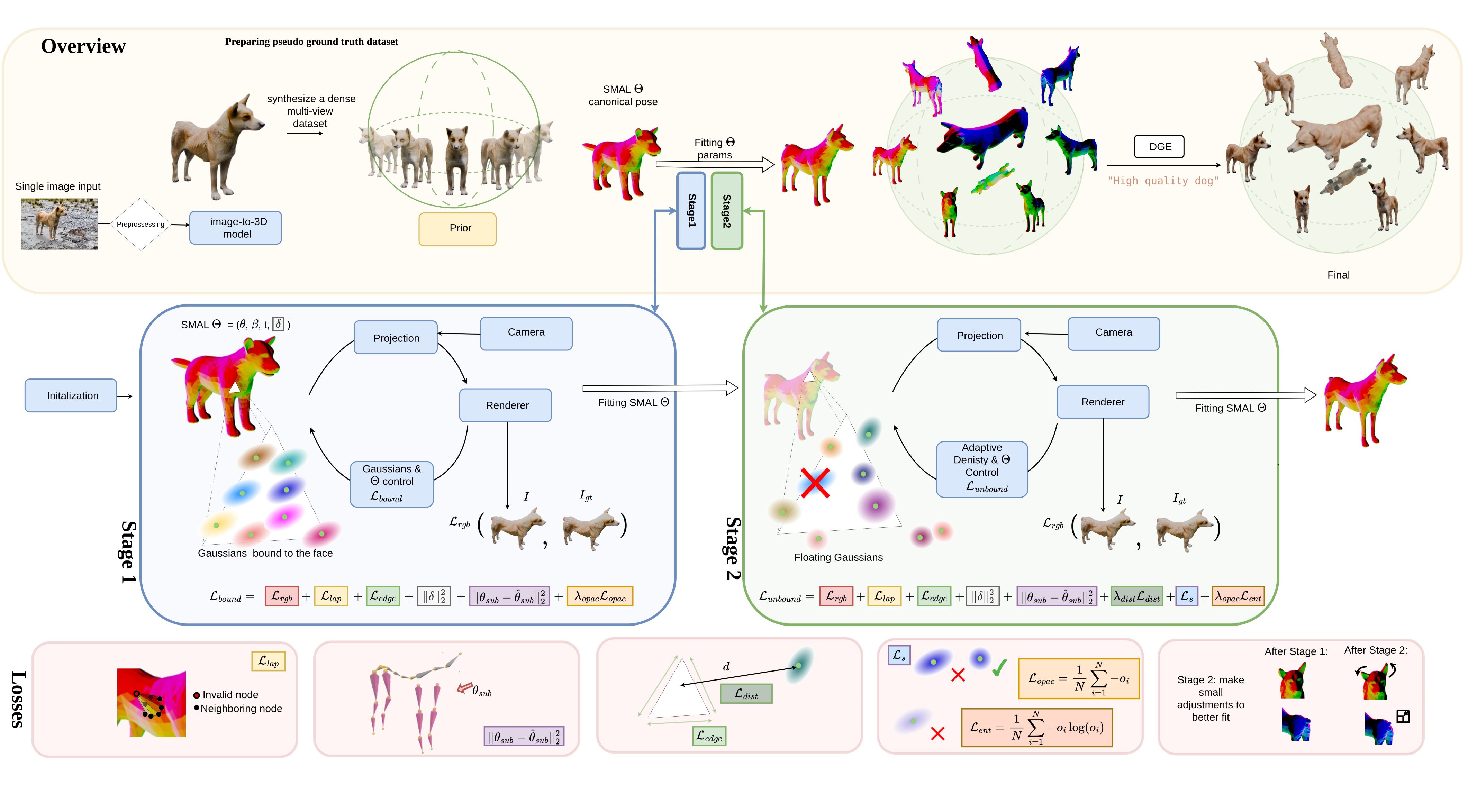}
\vspace{-1cm}
    \caption{Our model \our{} generate a dog avatar from a single input image by combining Gaussian primitives with a parametric SMAL mesh model. First, dense multi-view dataset is synthesised, which is considered pseudo ground truth. Stage 1: Gaussian primitives are bound to the faces of the SMAL model. Both the Gaussian and SMAL parameters are jointly optimized. Stage 2: Gaussians are allowed to detach from their associated faces, 
    adaptive density control alongside it is used to capture more details like fur. During this stage, the mesh itself is only slightly refined, while the primary improvements occur in the Gaussian representation.}
    \label{fig:model}
\vspace{-0.5cm}
\end{figure}

Furthermore, we address the animation bottleneck by enabling the avatar to execute complex behaviors through textual commands. By incorporating insights from motion synthesis and dynamic mesh modeling, such as FramePack~\cite{zhang2025frame} for temporal consistency and ActionMesh~\cite{sabathier2026actionmesh} for structural flexibility, or SMAL parameters. \our{} pipeline, inspired by the adaptability of Gaussian Splatting, provides a robust solution for artists and developers to generate, refine, and animate realistic animal avatars with minimal effort, see Fig.~\ref{fig:model}.

The primary contributions of this work are summarized as follows:

\vspace{-0.25cm}
\begin{enumerate}
    \item We propose a novel hybrid representation framework that integrates 3D Gaussian Splatting with the SMAL parametric model, effectively bridging the gap between high-fidelity visual representation and rigorous anatomical structure from only a single input image.
    \item We introduce a multimodal editing pipeline that leverages vision-language models to perform semantic refinement of animal avatars, allowing users to enhance surface textures, such as fur, through natural language prompts.
    \item We develop an automated motion-synthesis workflow that enables the animation of static animal reconstructions via textual commands, thereby bypassing manual rigging by translating generative video priors into actionable mesh deformations.
\end{enumerate}

\vspace{-0.6cm}
\section{Related Work}

\vspace{-0.25cm}
The development of \our{} sits at the intersection of several rapidly evolving fields in computer vision and computer graphics. To provide a comprehensive context for our work, we first review the current landscape of animal avatar creation and the limitations of existing species-specific models. We then examine the latest breakthroughs in single-image 3D reconstruction and the role of parametric models in providing anatomical grounding. Finally, we discuss the emergence of 3D Gaussian Splatting as a rendering primitive and explore recent advancements in text-driven editing and motion synthesis that enable the creation of dynamic, controllable digital models from static inputs.

\begin{wraptable}{r}{0.6\textwidth}
\vspace{-1.2cm}
\centering
{
\caption{Comparison of our \our{} method with methods that create dog avatars. In the task of creating an avatar based on a single image, only the DogRecon method has been published, but the code is not available. All these methods are based on the parametric SMAL dog mesh model.}
\centering
\setlength{\tabcolsep}{4.3pt}
{\fontsize{6.8pt}{11pt}\selectfont
\begin{tabular}{l|C{1.4cm}C{0.8cm}C{1cm}C{1.1cm}}
Methods & Dog \mbox{compatibility} & Single image & Animatable & Code available \\ \hline
AniAv.~\cite{sabathier2024animal} & \checkmark & & & \checkmark \\
GART~\cite{lei2024gart} & \checkmark &  & \checkmark & \checkmark \\
DogRecon~\cite{cho2025dogrecon} & \checkmark & \checkmark & \checkmark  \\
\textbf{\our{}} & \textbf{\checkmark }& \textbf{\checkmark }&\textbf{ \checkmark}& \checkmark \\
\end{tabular}
    \label{tab:models}}
    }
\vspace{-0.75cm}
\end{wraptable}

\noindent\textbf{Animal Avatar Creation} The field of animal reconstruction has evolved from simple shape estimation to complex avatar creation. Early foundational models including BARC~\cite{rueegg2022barc} and Common Pets in 3D~\cite{sinha2023common} established the feasibility of reconstructing species-specific geometry from limited data. Subsequent research focused on enhancing realism and diversity; MagicPony~\cite{wu2023magicpony} introduced unsupervised learning for articulated shapes, while Generative Zoo~\cite{niewiadomski2025generative} utilized generative priors to handle the vast morphological variety of animals. For high-fidelity results, DogRecon~\cite{cho2025dogrecon} and DogMo~\cite{wang2025dogmo} utilize neural representations to capture intricate details. To address dynamics, AnimalAvatar (AniAv.)~\cite{sabathier2024animal}, GART~\cite{lei2024gart}, and 4D-Animal~\cite{zhong20254d} extend these models into the temporal domain, though they often require multi-view data or video sequences for stable results, see Tab.~\ref{tab:models}.

\noindent\textbf{Single-Image 3D Reconstruction}
Generating 3D assets from a single viewpoint is an ill-posed problem that has recently seen a breakthrough via large-scale feed-forward models. SAM3D~\cite{yang2023sam3d} leverages segment-anything priors to isolate geometry, while TripoSG~\cite{li2025triposg} and Hunyuan 3D~\cite{yang2024hunyuan3d} offer rapid mesh generation from single images. More recently, TRELLIS~\cite{xiang2025structured} has demonstrated superior structural quality by utilizing structured latent spaces. While these tools excel at generating static geometry, the resulting meshes often lack the anatomical rigging required for animation.

\begin{figure}[t]
\centering
\setlength{\tabcolsep}{1pt}
{\fontsize{6.8pt}{11pt}\selectfont
\begin{NiceTabular}{l ccr c ccr c ccr c}
 Input & \multicolumn{3}{c}{TripoSG}  & \hspace{2mm} &
  \multicolumn{3}{c}{SAM3D}  & \hspace{2mm} &
  \multicolumn{3}{c}{Trellis} & \\
\Block{2-1}{\includegraphics[width=0.085\linewidth, trim={50, 0, 50, 0}, clip]{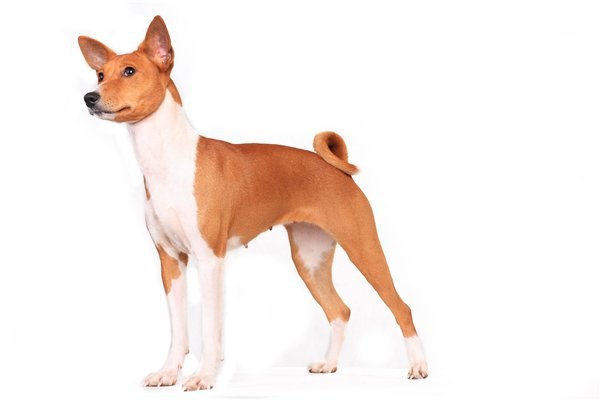}} & 
\includegraphics[height=0.085\linewidth, trim={50, 50, 50, 100}, clip]{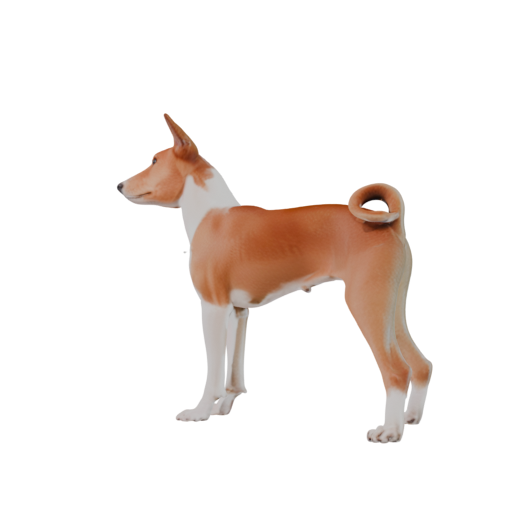} & 
\includegraphics[height=0.085\linewidth, trim={50, 50, 125, 100}, clip]{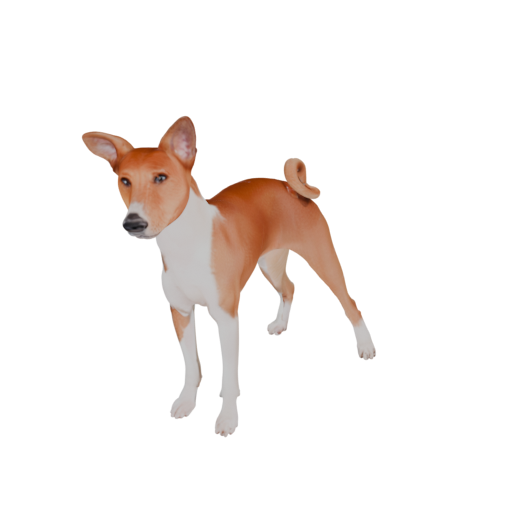} & 
\includegraphics[height=0.085\linewidth, trim={50, 250, 300, 100}, clip]{imgs/triposg/dog6/00048.png} & &
\includegraphics[height=0.085\linewidth, trim={50, 75, 50, 100}, clip]{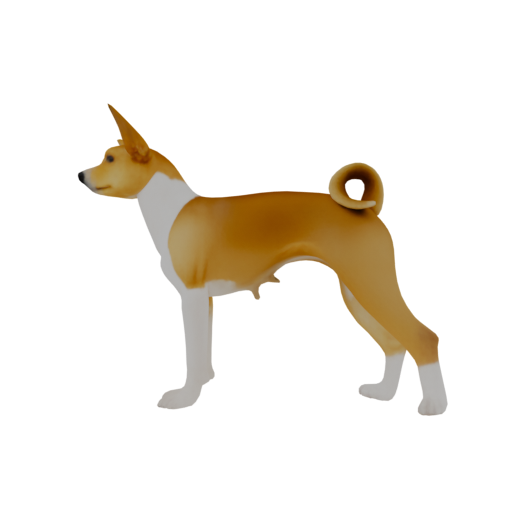} & 
\includegraphics[height=0.085\linewidth, trim={50, 75, 150, 100}, clip]{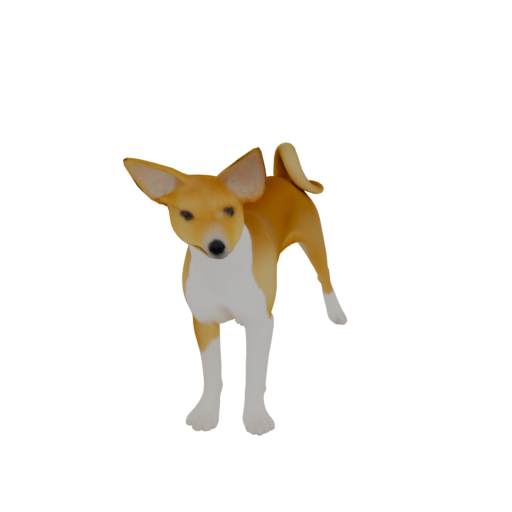} & 
\includegraphics[height=0.085\linewidth, trim={120, 200, 240, 125}, clip]{imgs/sam3d/dog6/00048.png} & &
\includegraphics[height=0.085\linewidth, trim={50, 75, 50, 100}, clip]{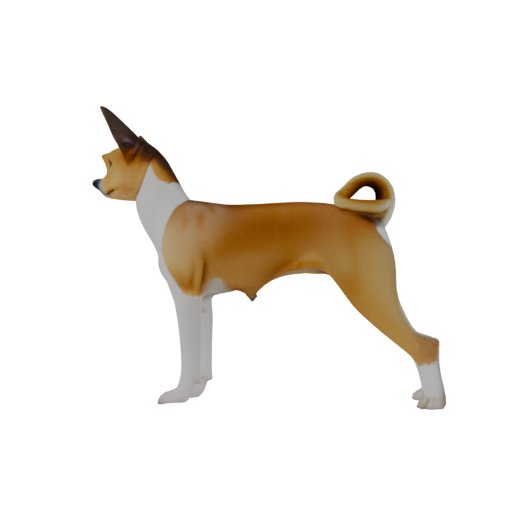} & 
\includegraphics[height=0.085\linewidth, trim={50, 75, 150, 100}, clip]{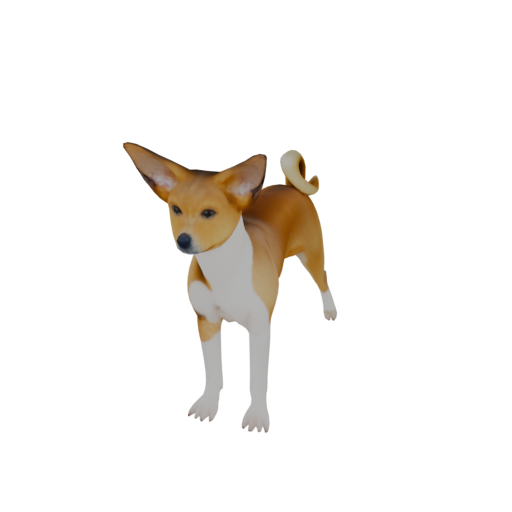} & 
\includegraphics[height=0.085\linewidth, trim={110, 230, 240, 110}, clip]{imgs/trellis/dog6/00048.png} &
\raisebox{3.5em}{\rotatebox{-90}{{Baseline}} } \\
&
\Block[draw=[RGB]{0,153,76}, rounded-corners]{1-11}{}
\includegraphics[height=0.085\linewidth, trim={50, 50, 50, 100}, clip]{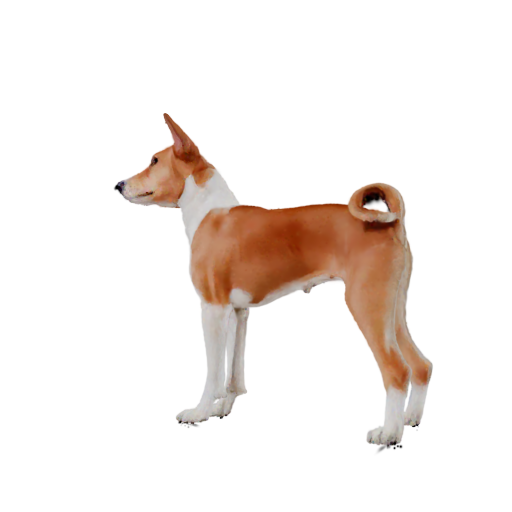} & 
\includegraphics[height=0.085\linewidth, trim={50, 50, 125, 100}, clip]{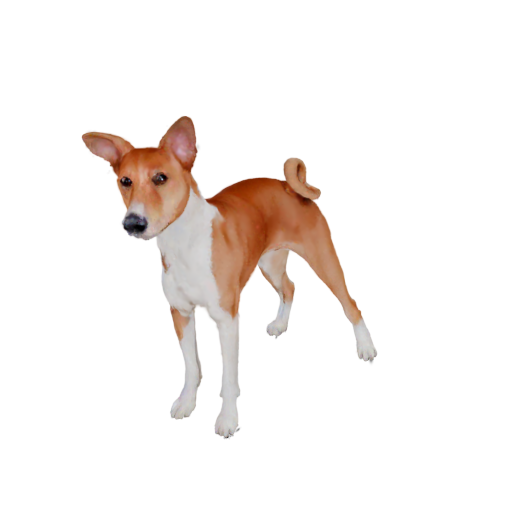} & 
\includegraphics[height=0.085\linewidth, trim={50, 250, 300, 100}, clip]{imgs/triposg/dog6_our/00048.png} & &
\includegraphics[height=0.085\linewidth, trim={50, 75, 50, 100}, clip]{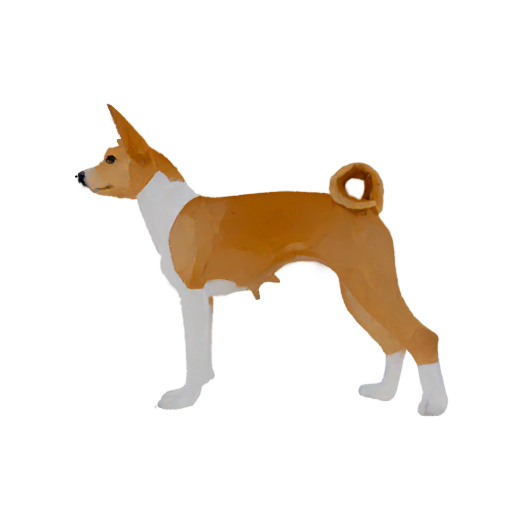} & 
\includegraphics[height=0.085\linewidth, trim={50, 75, 150, 100}, clip]{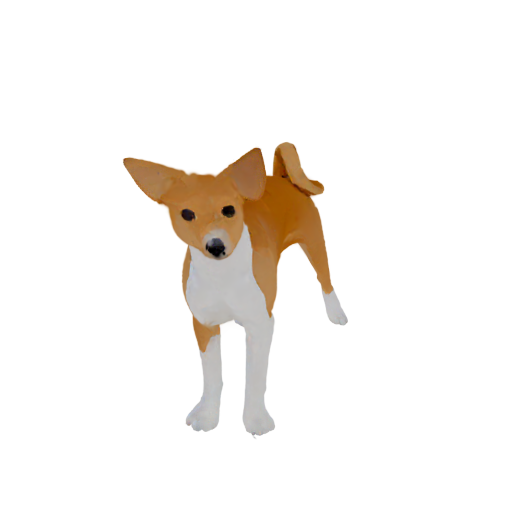} & 
\includegraphics[height=0.085\linewidth, trim={120, 200, 240, 125}, clip]{imgs/sam3d/dog6_our/00048.png} & &
\includegraphics[height=0.085\linewidth, trim={50, 75, 50, 100}, clip]{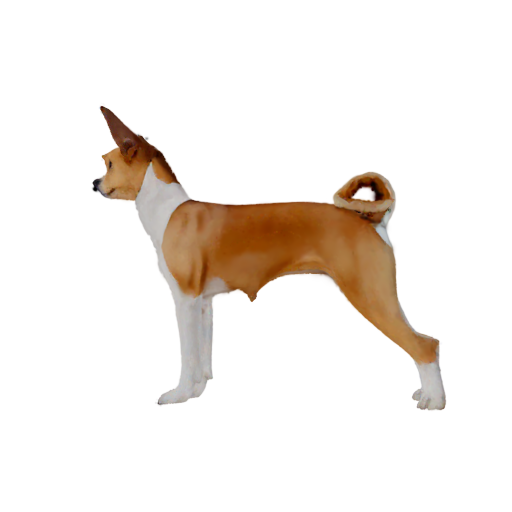} & 
\includegraphics[height=0.085\linewidth, trim={50, 75, 150, 100}, clip]{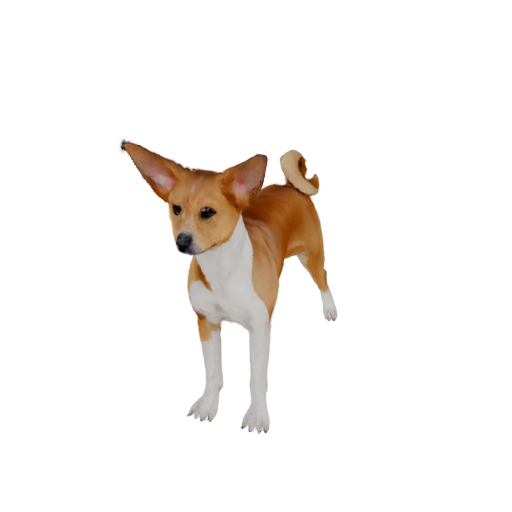} & 
\includegraphics[height=0.085\linewidth, trim={110, 230, 240, 110}, clip]{imgs/trellis/dog6_our/00048.png} &
\raisebox{2.7em}{\rotatebox{-90}{Ours}} \\

\noalign{\vskip 1mm} 
\Block{2-1}{
\includegraphics[height=0.085\linewidth, trim={50, 0, 50, 0}, clip]{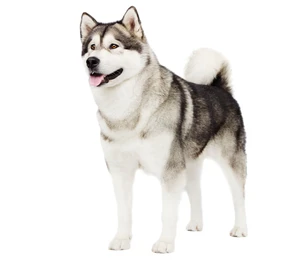}
} &
\includegraphics[height=0.085\linewidth, trim={50, 100, 50, 100}, clip]{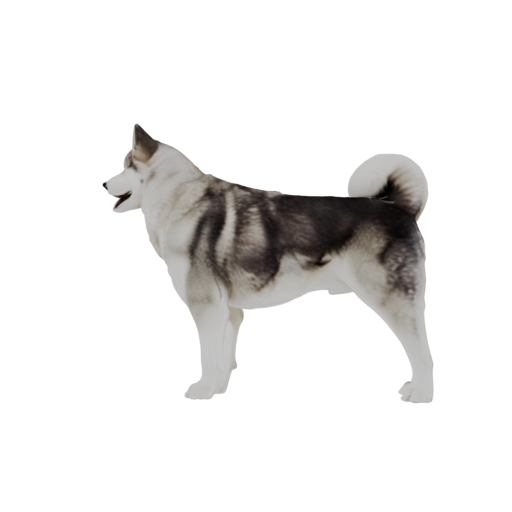} & 
\includegraphics[height=0.085\linewidth, trim={150, 100, 150, 100}, clip]{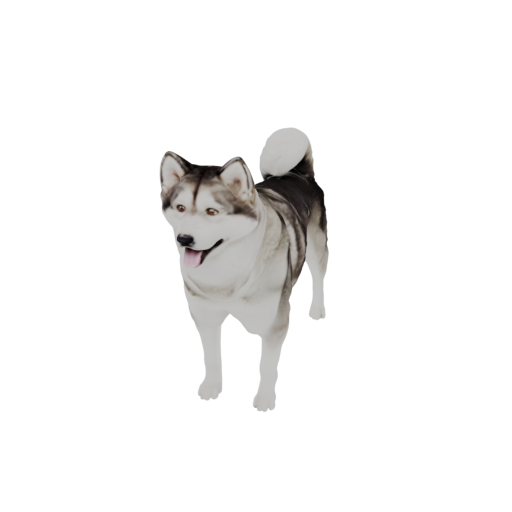} & 
\includegraphics[height=0.085\linewidth, trim={150, 200, 260, 150}, clip]{imgs/triposg/dog7/00048.png} & &
\includegraphics[height=0.085\linewidth, trim={20, 50, 50, 70}, clip]{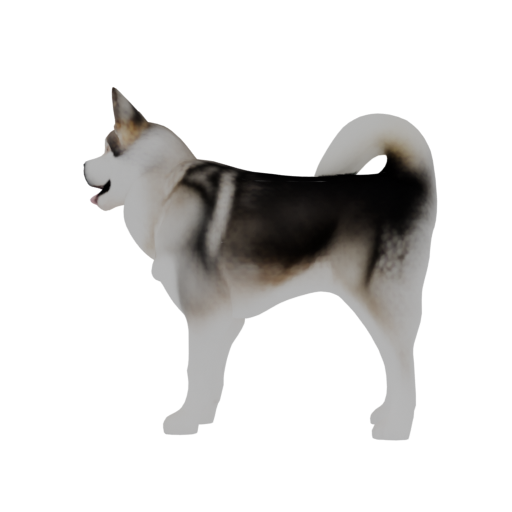} & 
\includegraphics[height=0.085\linewidth, trim={100, 50, 150, 70}, clip]{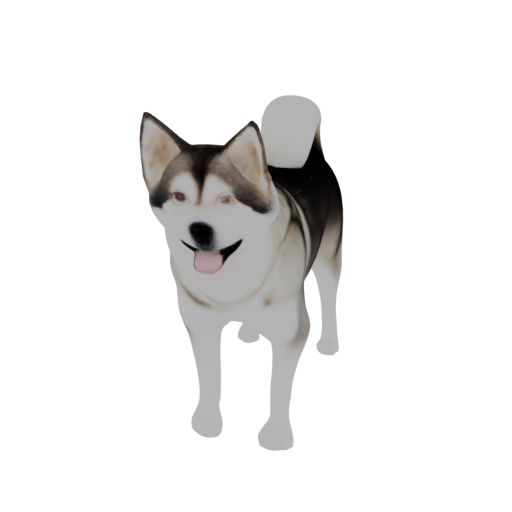} & 
\includegraphics[height=0.085\linewidth, trim={110, 200, 240, 100}, clip]{imgs/sam3d/dog7/00048.png} & &
\includegraphics[height=0.085\linewidth, trim={50, 100, 50, 100}, clip]{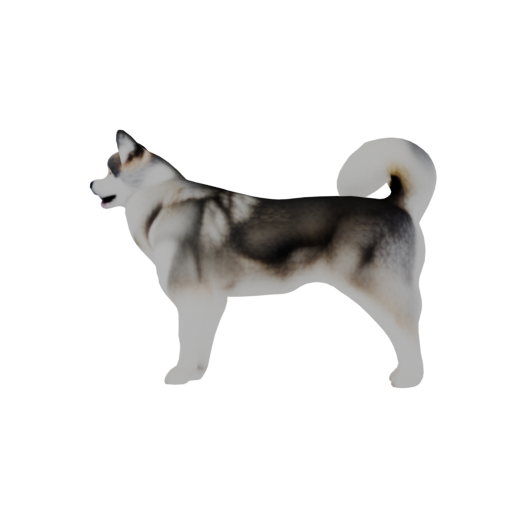} & 
\includegraphics[height=0.085\linewidth, trim={150, 100, 150, 100}, clip]{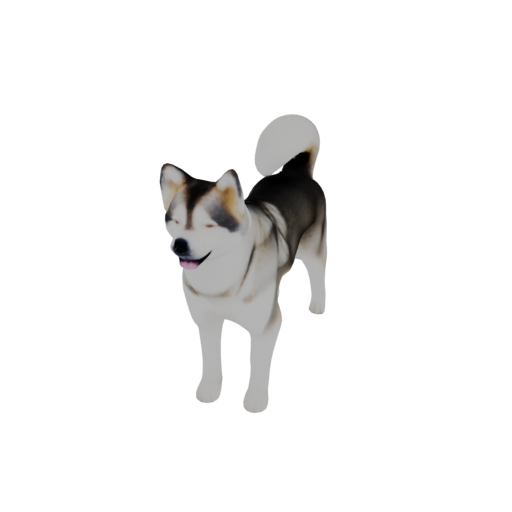} & 
\includegraphics[height=0.085\linewidth, trim={150, 200, 260, 150}, clip]{imgs/trellis/dog7/00048.png} &
\raisebox{3.5em}{\rotatebox{-90}{Baseline}} \\
&
\Block[draw=[RGB]{0,153,76}, rounded-corners]{1-11}{}
\includegraphics[height=0.085\linewidth, trim={50, 100, 50, 100}, clip]{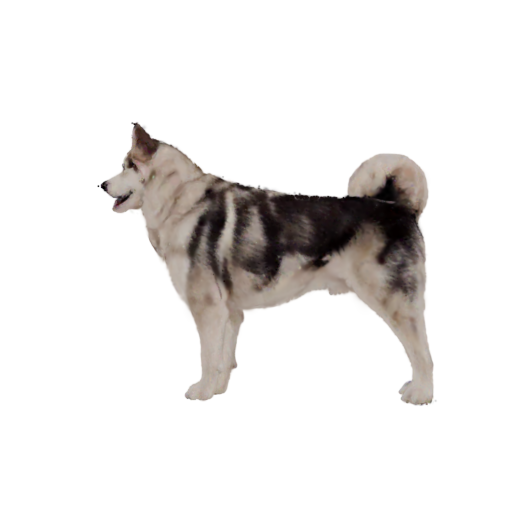} & 
\includegraphics[height=0.085\linewidth, trim={150, 100, 150, 100}, clip]{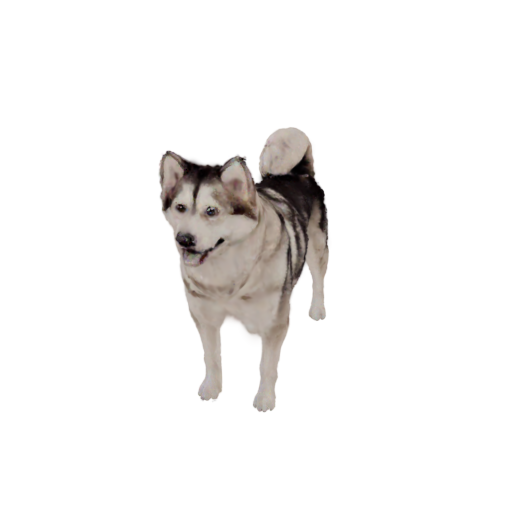} & 
\includegraphics[height=0.085\linewidth, trim={150, 200, 260, 150}, clip]{imgs/triposg/dog7_our/00048.png} & &
\includegraphics[height=0.085\linewidth, trim={20, 50, 50, 70}, clip]{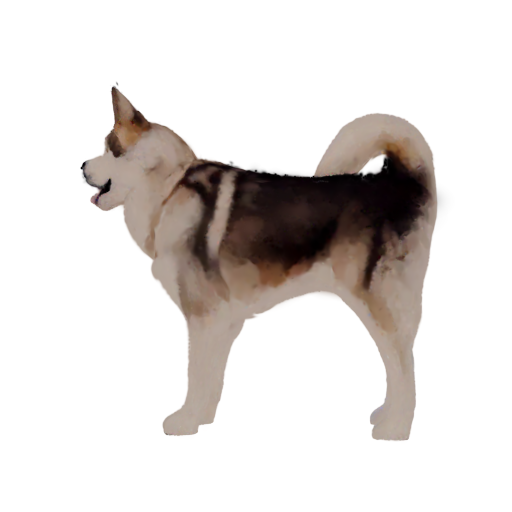} & 
\includegraphics[height=0.085\linewidth, trim={100, 50, 150, 70}, clip]{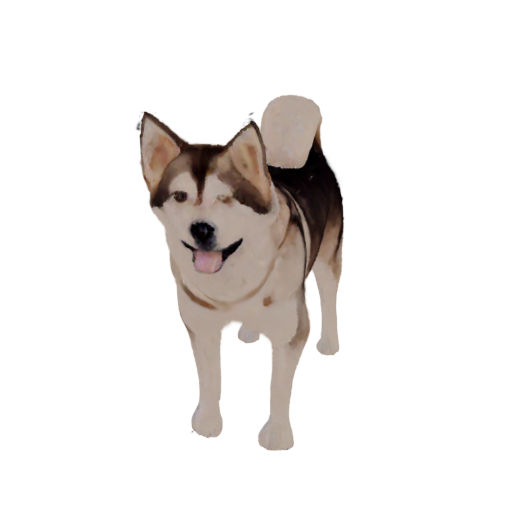} & 
\includegraphics[height=0.085\linewidth, trim={110, 200, 240, 100}, clip]{imgs/sam3d/dog7_our/00048.png} & &
\includegraphics[height=0.085\linewidth, trim={50, 100, 50, 100}, clip]{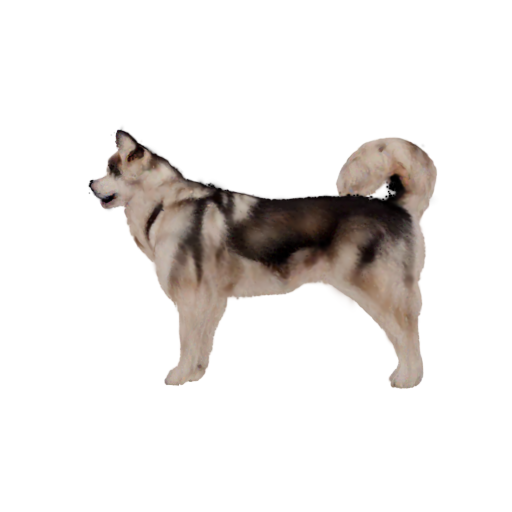} & 
\includegraphics[height=0.085\linewidth, trim={150, 100, 150, 100}, clip]{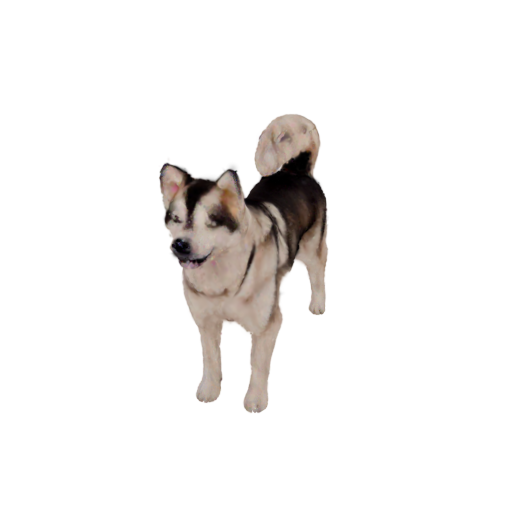} &
\includegraphics[height=0.085\linewidth, trim={150, 200, 260, 150}, clip]{imgs/trellis/dog7_our/00048.png} &
\raisebox{2.7em}{\rotatebox{-90}{Ours}}

\end{NiceTabular}
}
\caption{To construct a synthetic dataset representing pseudo ground truth, we apply our method to firstly generated 3D assets by three state-of-the-art image-to-3D approaches: TripoSG, SAM3D, and Trellis. Since our model uses Gaussian primitives, we can faithfully reproduce the geometry and appearance of these generated assets while maintaining structural consistency.  Furthermore, by leveraging Gaussian editing-based methods like DGE,  dog avatar can be improved in particularity quality of high-frequency details, mitigating the common ``plastic'' appearance observed in firstly generated assets.}
\label{fig:inputdataset}
\vspace{-0.6cm}
\end{figure}

\noindent\textbf{Parametric Mesh Fitting}
To transform raw geometry into a functional avatar, parametric models provide necessary structural constraints. The SMAL model~\cite{zuffi20173d} remains the gold standard for articulated animal shapes, offering a biologically inspired linear space for various mammals. Recent improvements like BITE~\cite{ruegg2023bite} refine the fitting process by incorporating behavioral and pose priors, ensuring that the mesh remains plausible even under extreme deformations. Our work builds on these fitting techniques to align the SMAL topology with unstructured Gaussian data.

\noindent\textbf{Gaussian Splatting and Text-Driven Editing}
3D Gaussian Splatting (3DGS) has revolutionized real-time rendering, but editing these representations remains a challenge. GaussianEditor~\cite{wang2024gaussianeditor} proposed a swift and controllable 3D editing framework for Gaussian splatting, enabling efficient semantic and geometric manipulations with intuitive control, laying a foundational basis for text-driven GS editing. DGE~\cite{chen2024dge} and Morpheus~\cite{wynn2025morpheus} introduce natural language interfaces for manipulating GS scenes, allowing for semantic changes in texture and geometry. These tools are essential for overcoming the ``plastic look'' of initial reconstructions by refining surface details through text-based guidance.

\begin{wraptable}{r}{0.55\textwidth}
\vspace{-1.2cm}
\caption{Comparison of different backbones using CLIP-Score between input image and our model renders averaged over all test views after 2nd stage and Final.}
\centering
\setlength{\tabcolsep}{4.3pt}
{\fontsize{6.8pt}{11pt}\selectfont{
\begin{NiceTabular}{lcccl}
         & \multicolumn{3}{c}{CLIP-Score} \\  \cmidrule(lr){2-4} 
         \hspace{-2mm}Backbones& SAM3D & Trellis & TripoSG & after\\ \hline{1-4}
        \Block{2-1}{\includegraphics[width=0.07\textwidth]{imgs/inputs/dog6.jpg}} & 0.815 & 0.829 & \textbf{0.860}  & stage 2\\
        & \textbf{0.837} & \textbf{0.863} & \textbf{0.886} & Final\\ \hdashline
        \Block{2-1}{\includegraphics[width=0.07\textwidth]{imgs/inputs/dog7.png}} & 0.813 & 0.818 & \textbf{0.852}  & stage 2\\
        & \textbf{0.843 }& \textbf{0.858} & \textbf{0.858} & Final\\
\end{NiceTabular}
}\label{tab:backbones}}
\vspace{-0.9cm}
\end{wraptable}

\noindent\textbf{Dynamic Meshes and Motion Synthesis}
Animating static reconstructions requires mapping motion onto a deformable structure. Artemis~\cite{luo2022artemis} and Make-It-Animatable~\cite{guo2025make} focus on creating rigged systems from static inputs. MagicArticulate~\cite{song2025magicarticulate} enables robust articulation and rigging for general articulated objects from sparse observations, strengthening the pipeline for animating static 3D content.  For animal-specific motion, ActionMesh~\cite{sabathier2026actionmesh} and AnimalGS~\cite{li2026animalgs} provide frameworks for realistic deformation. Integrating temporal consistency from single images is further supported by FramePack~\cite{zhang2025frame}, which aids in synthesizing video-driven motion. Furthermore, recent research addresses the challenge of long-range controllable dynamics. TC4D~\cite{tc4d} achieves this by decoupling the motion into global trajectories and local transformations, while 3DTrajMaster~\cite{fu20243dtrajmaster} injects control into video generation process, enabling precise trajectory-aware manipulation. Finally, methods like Gaussian See, Gaussian Do~\cite{bekor2025gaussian} and GaussianMorphing~\cite{li2025gaussianmorphing} explore the synergy between motion transfer and GS representations, providing the basis for our text-to-command animation pipeline.

\section{Method}

\begin{wrapfigure}{r}{0.55\textwidth}
\vspace{-1.5cm}
 \centering
\setlength{\tabcolsep}{4pt}
{\fontsize{6.8pt}{11pt}\selectfont
\begin{tabular}{ccc}
Input image & \hspace{0.3cm} DogRecon \hspace{0.3cm} & \hspace{0.3cm}\our{} (our)\\
\end{tabular}
  \includegraphics[width=0.55\textwidth, trim={0, 50
  , 0, 10}, clip]{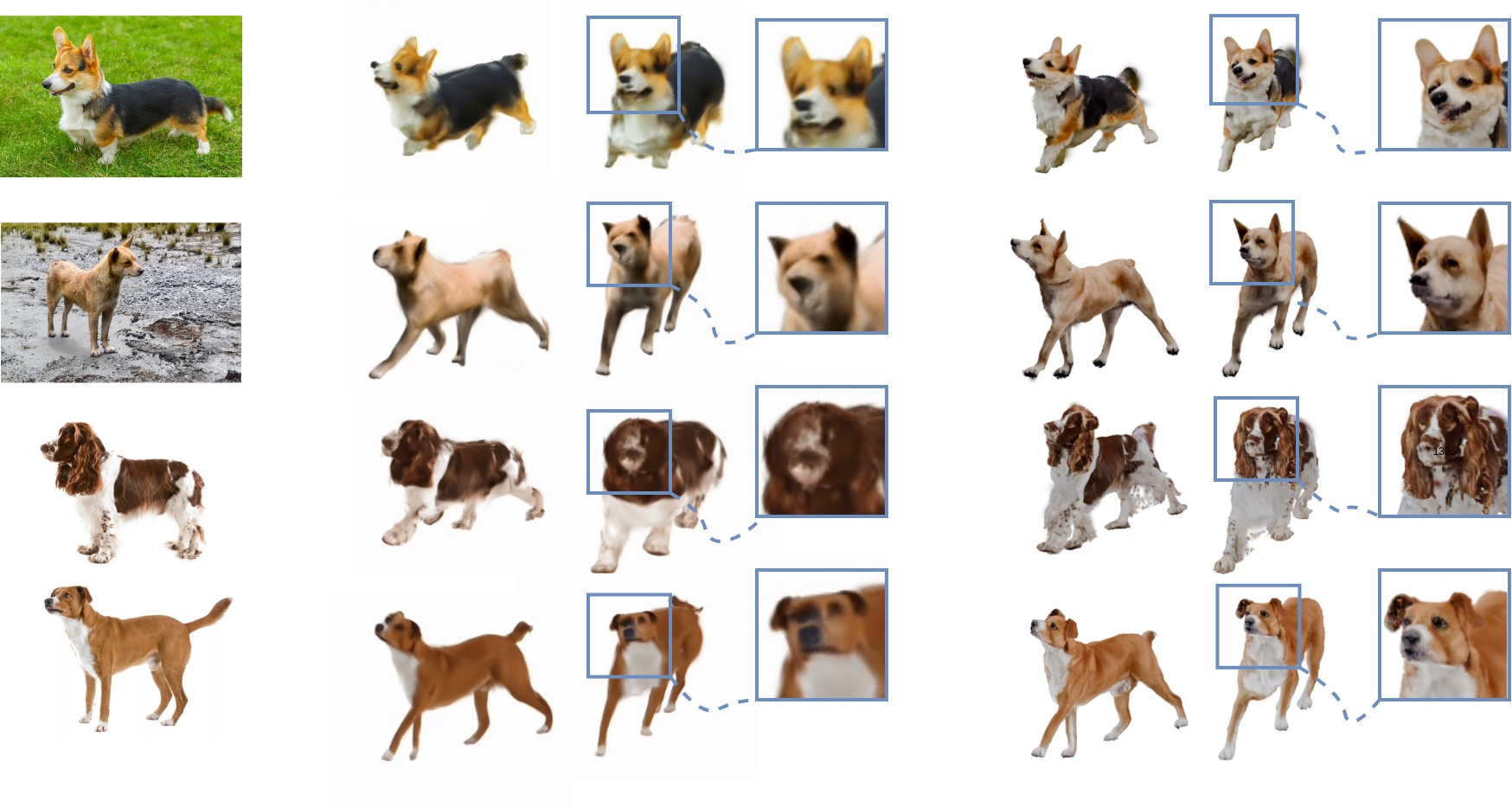}
}
\vspace{-0.65cm}
\caption{Visual comparison with state-of-the-art method DogRecon for single-image dog avatar reconstruction. Both approaches represent the avatar using 3D Gaussians combined with the parametric SMAL model. Our method shows improved details. Additionally, the SMAL representation enables animation through direct manipulation of parameters.}
\label{fig:comparisionwithdocracon}
\vspace{-0.8cm}
\end{wrapfigure}

Our method introduces a framework to reconstruct and animate 3D animal avatars from a single RGB image by combining the high-fidelity volumetric rendering of 3DGS with the geometric and structural priors of the SMAL model. The pipeline begins with generating a synthetic multi-view dataset from the single input image. The reconstruction process is split into two distinct stages, a mesh-bound initial optimization  and an unbound volumetric refinement. The reconstruction is followed by a skinning process that rigs the Gaussian splats to the underlying mesh for novel poses animation. Fig.~\ref{fig:model} shows an overview of our pipeline.

\vspace{-0.5cm}
\subsection{Preparation for training}
\vspace{-0.25cm}

To construct a dog avatar, we assume a well-exposed dog input image. Contemporary state-of-the-art methods significant degraded under non-canonical poses. Therefore, we recommend an optional normalization step that edits the input toward a canonical configuration (e.g. view from the side, without large deviations from the standing position, tail well presented) via prompt-guided image transformation. We find both Qwen-Image-Edit-2511~\cite{wu2025qwenimagetechnicalreport} and GPT-4o~\cite{gpt4o} effective for this task. This step can be omitted when the input already close to the canonical pose.

\begin{wrapfigure}{r}{0.55\textwidth}
\vspace{-1cm}
 \centering
\setlength{\tabcolsep}{0.6cm}
{\fontsize{6.8pt}{11pt}\selectfont
\begin{tabular}{ccc}
GT & \hspace{3mm}DogRecon & \hspace{-4mm}\our{} (Our)
\end{tabular}}
    \includegraphics[width=\linewidth, trim={0, 60, 0, 0}, clip]{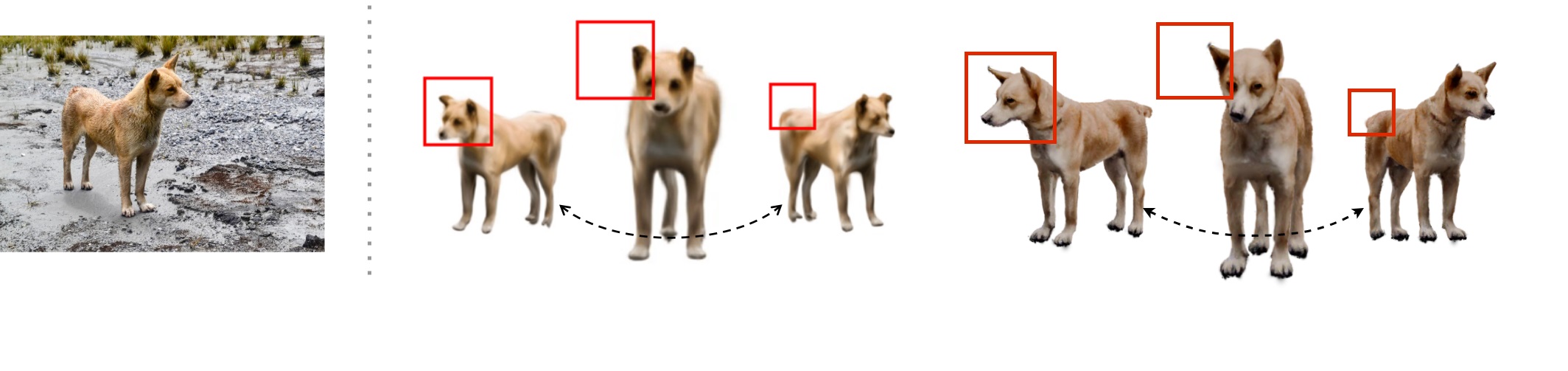}
\caption{Visual comparison with DogRecon from three viewpoints. Our method preserves more details appearance, resulting in a more realistic rendering. }
\label{fig:dogreconvsour}
\vspace{-0.9cm}
\end{wrapfigure}

We represent the avatar as a hybrid of a parametric mesh and a collection of Gaussian splats. The underlying structural prior is defined by the SMAL model, which we consider as a differentiable function $M(\cdot)$ that transforms parameters into mesh. The vertex positions $\mathcal{V} \in \mathbb{R}^{3889 \times 3}$ are computed as:
$
   \mathcal{V} = M(\Theta) = M(\beta, \theta, \mathbf{t}, \mathbf{d}),
$
where $\beta$ and $\theta$ represent the shape and pose coefficients, $\mathbf{t}$ is the global translation, and $\mathbf{d}$ is a set of learnable vertex offsets used to capture specific animal morphologies. It is initialized in the canonical position and all of its parameters $\Theta$ are trained during optimization.

\setlength\dashlinedash{0.2pt}
\begin{figure*}[t]
 \centering
\setlength{\tabcolsep}{4pt}
{\fontsize{6.8pt}{11pt}\selectfont
\begin{tabular}{cc@{\;}c@{}c@{}c@{}c@{}c@{}}
& Dataset & Preproc
& Our & DogRecon & AniAv. & GART \\ 
 \hline
  \rotatebox{90}{\hspace{-1cm}\fontsize{6.8pt}{11pt}{GART dataset}} &  \includegraphics[width=1.5cm,  trim={440, 0, 300, 0}, clip]{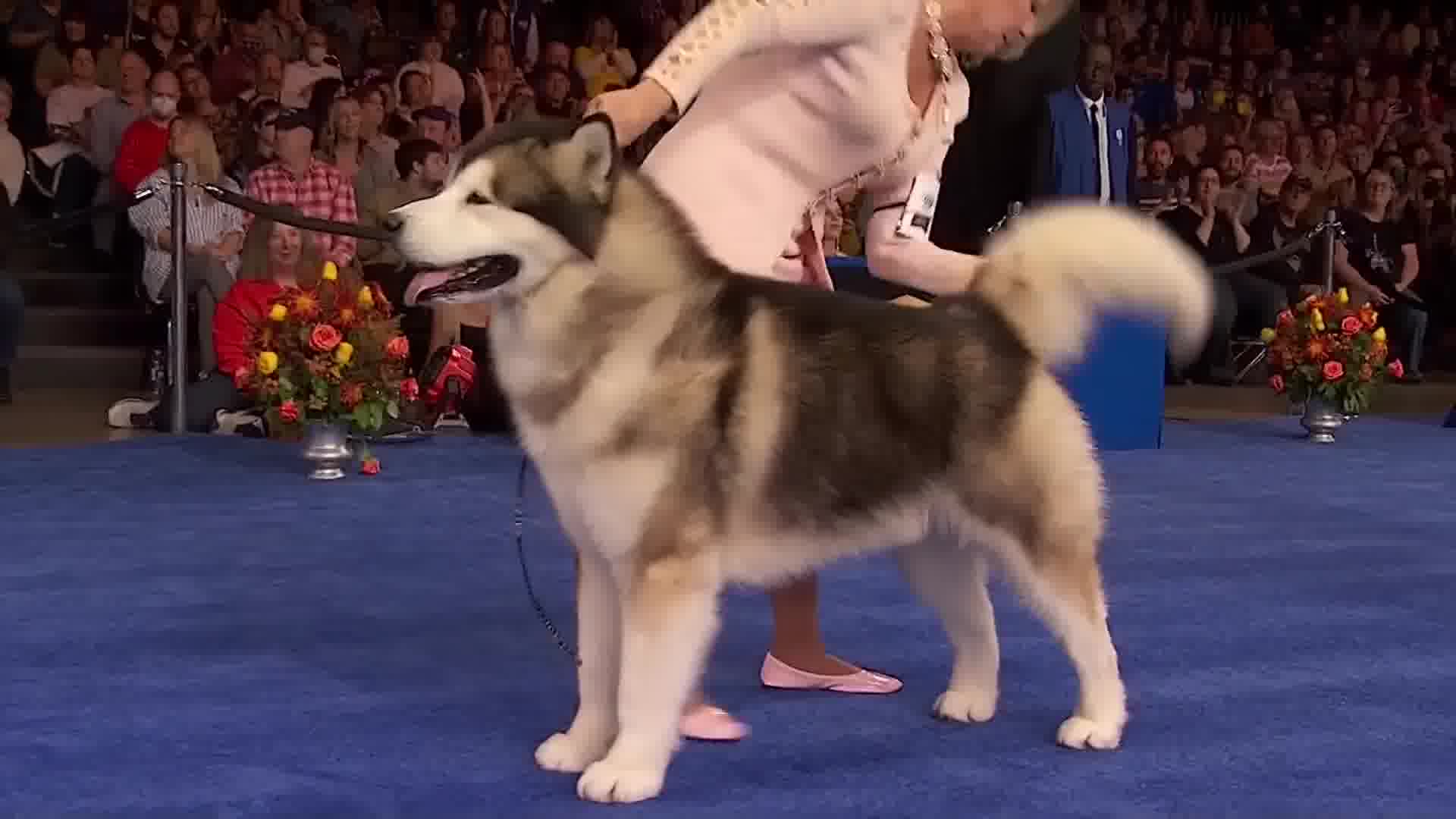} &
 \includegraphics[width=1.5cm,  trim={170, 0, 50, 0}, clip]{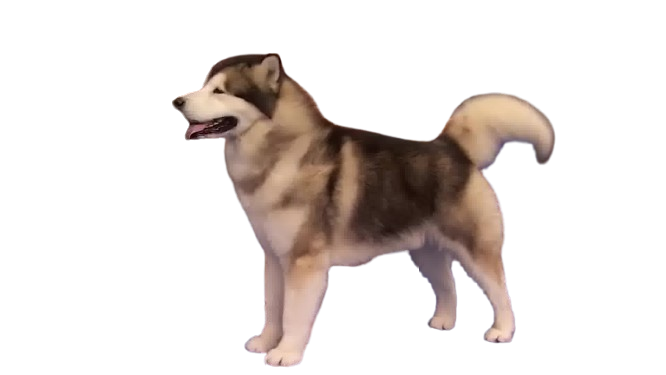} &
 \includegraphics[height=1.5cm, trim={60, 80, 120, 90}, clip]{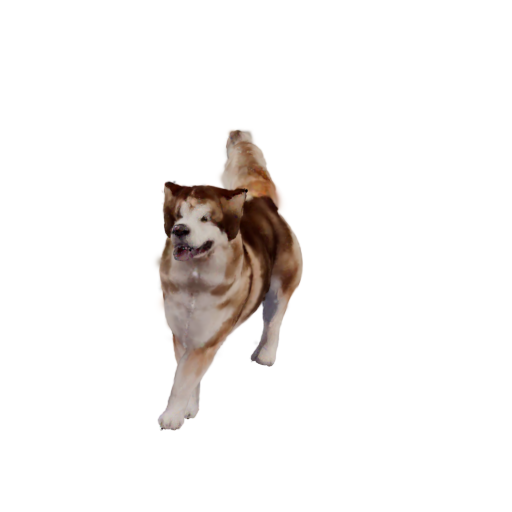}  & 
 \includegraphics[height=1.5cm]{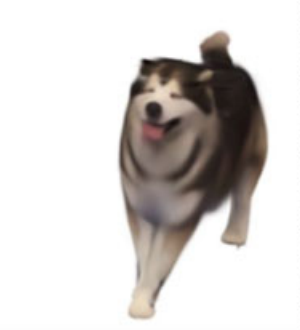} &
oom &
   \includegraphics[width=1.5cm, trim={110, 120, 110, 130}, clip]{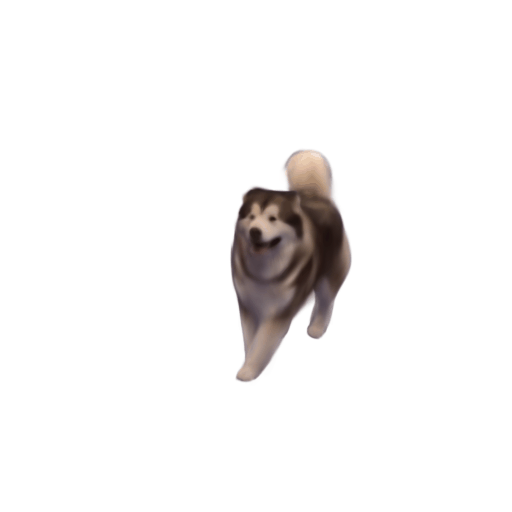} \\
 & \includegraphics[width=1.5cm, trim={0, 0, 0, 0}, clip]{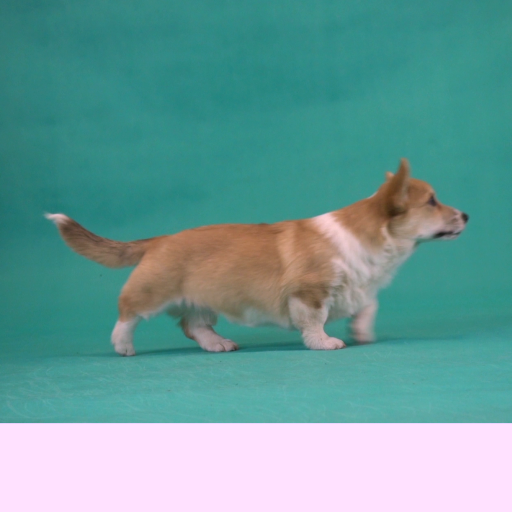} &
    \includegraphics[width=1.5cm, trim={150, 0, 150, 0}, clip]{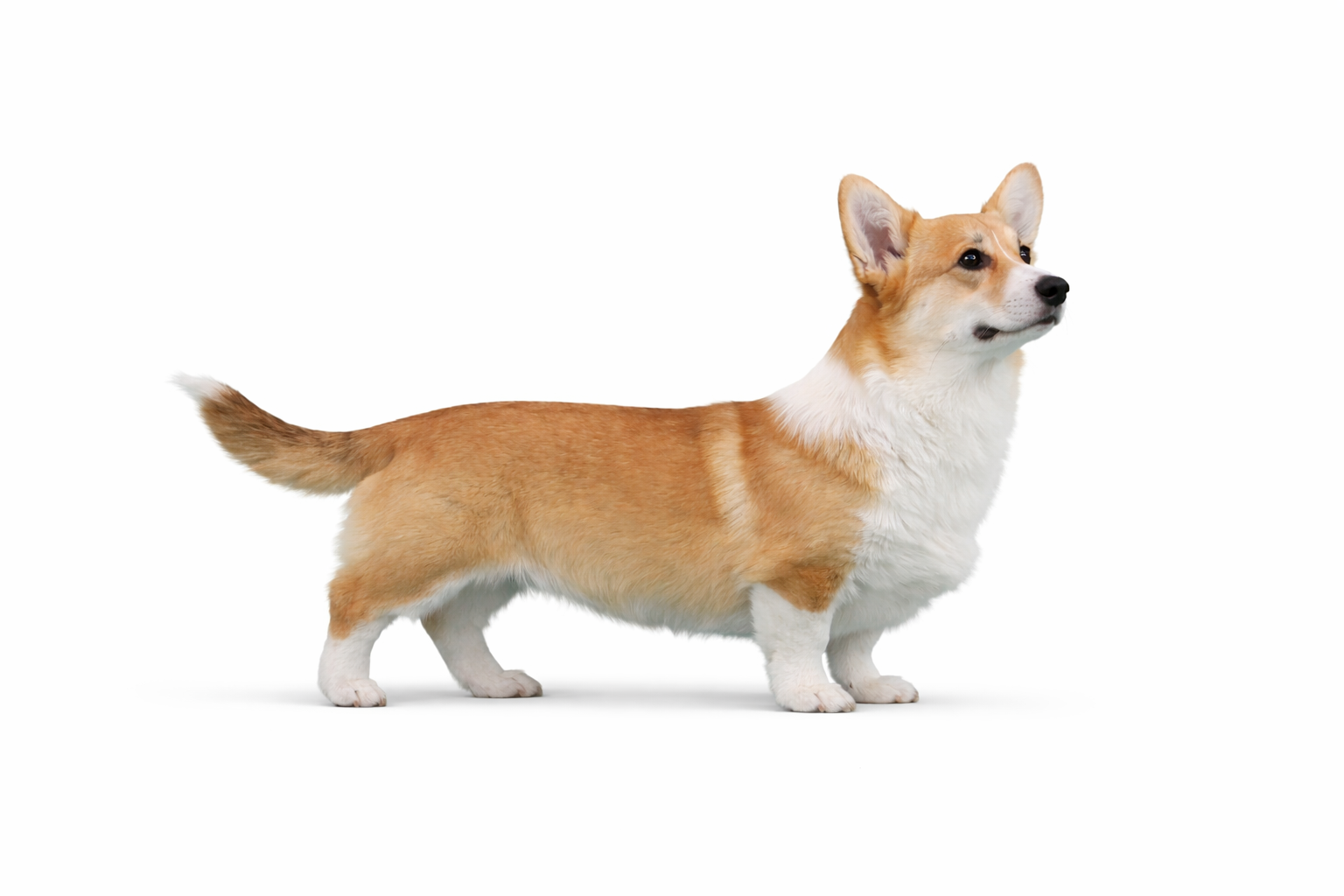} &
 \includegraphics[width=1.5cm, trim={10, 140, 180, 120}, clip]{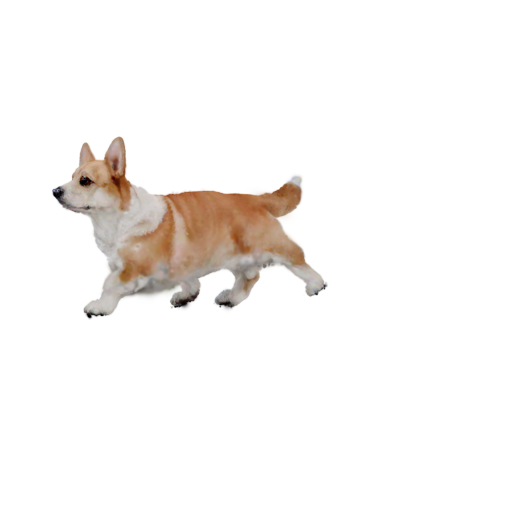} & 
 \includegraphics[width=1.5cm]{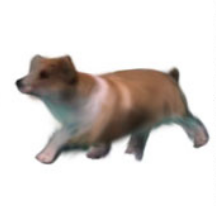} & &
   \includegraphics[width=1.5cm, trim={90, 110, 100, 100}, clip]{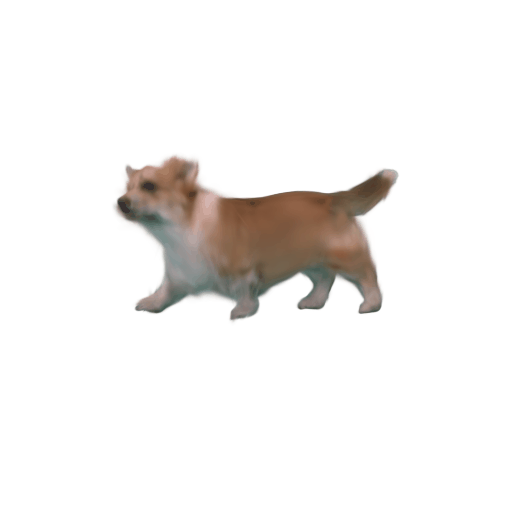} \\
    \hdashline
 \rotatebox{90}{\hspace{-0.8cm}\fontsize{6.8pt}{11pt}{CoP3D}} & \includegraphics[width=1.5cm,  trim={0, 800, 0, 33}, clip]{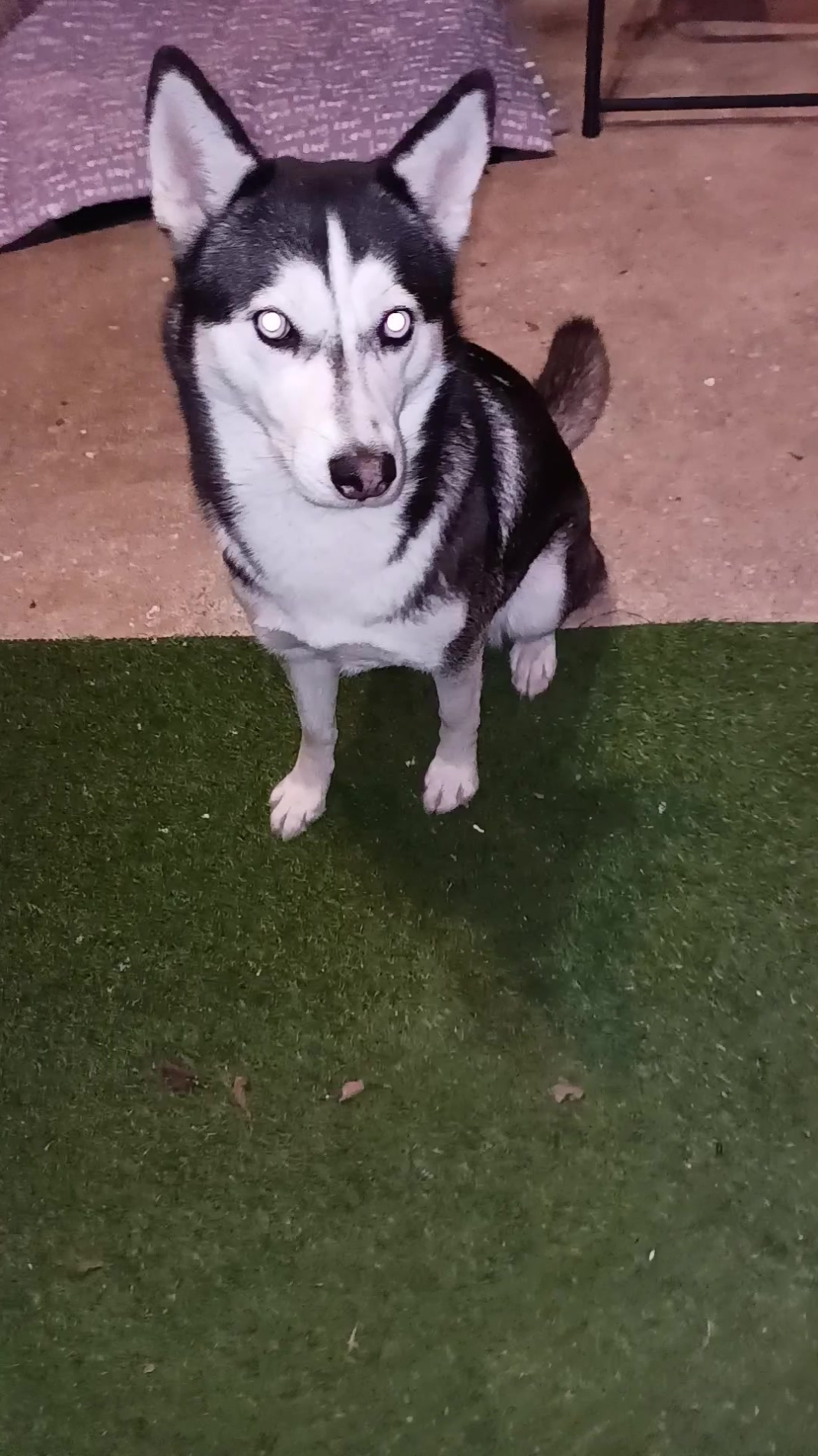} &
 \includegraphics[height=1.5cm, trim={70, 0, 110, 0}, clip]{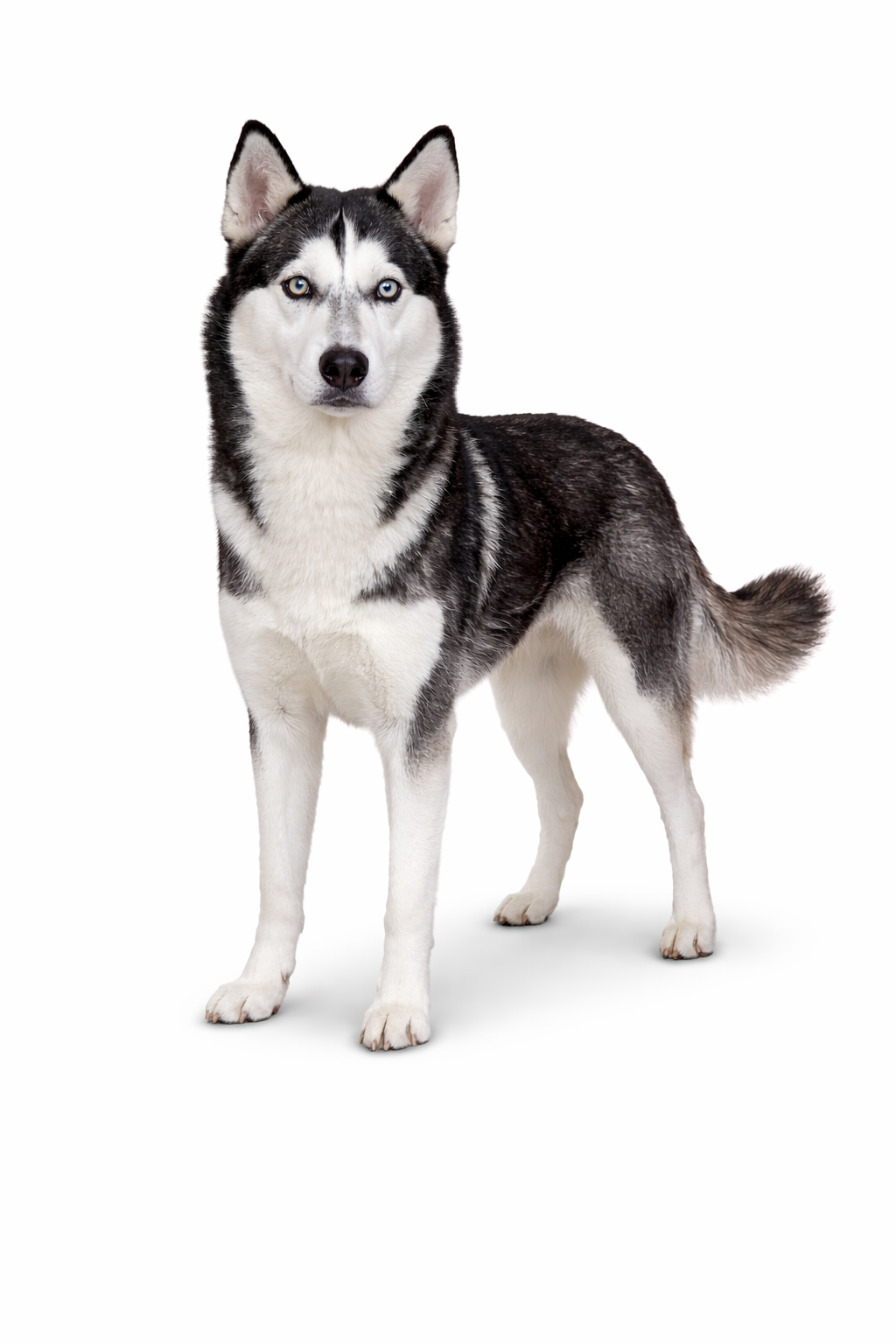} &
 \includegraphics[width=1.5cm, trim={110, 150, 90, 50}, clip]{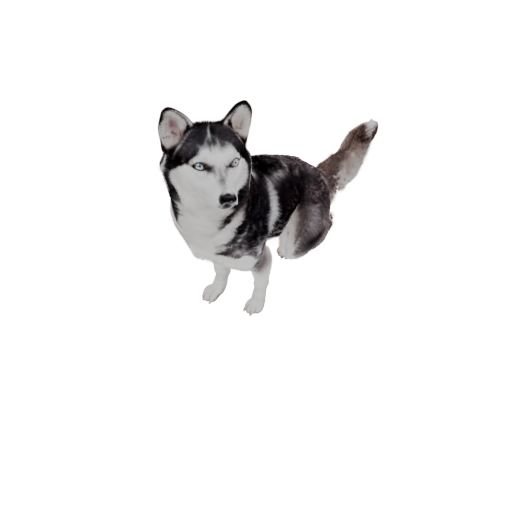} & 
{no}&
  \includegraphics[width=1.5cm, trim={0, 70, 100, 0}, clip]{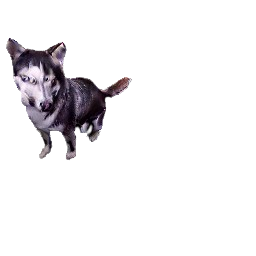} &
   \includegraphics[width=1.5cm, trim={90, 140, 100, 50}, clip]{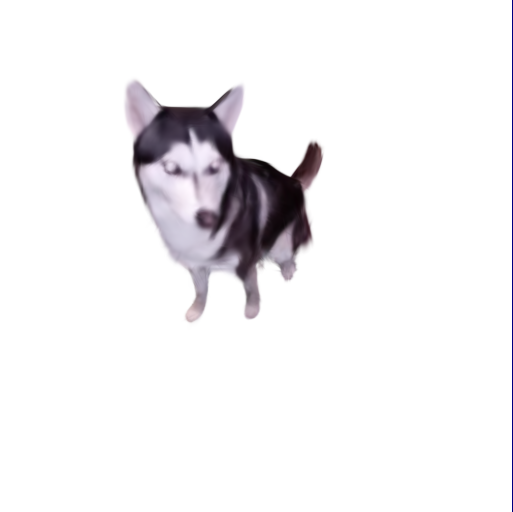}  \\
     &  \includegraphics[height=1.5cm]{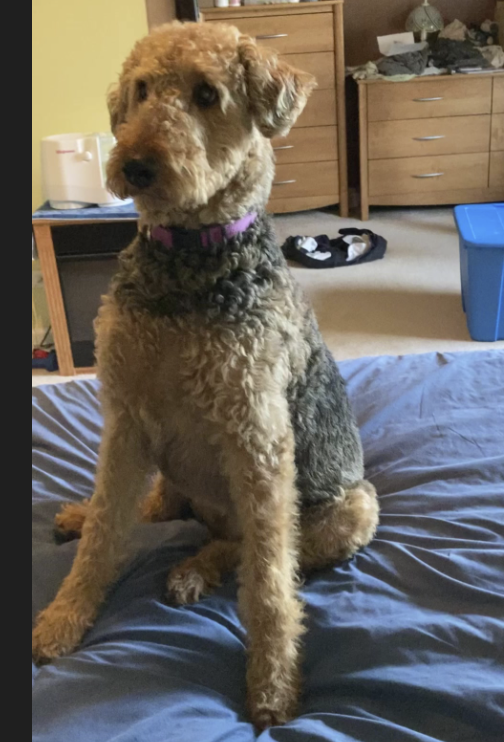} & 
 \includegraphics[width=1.5cm, trim={200, 0, 200, 100}, clip]{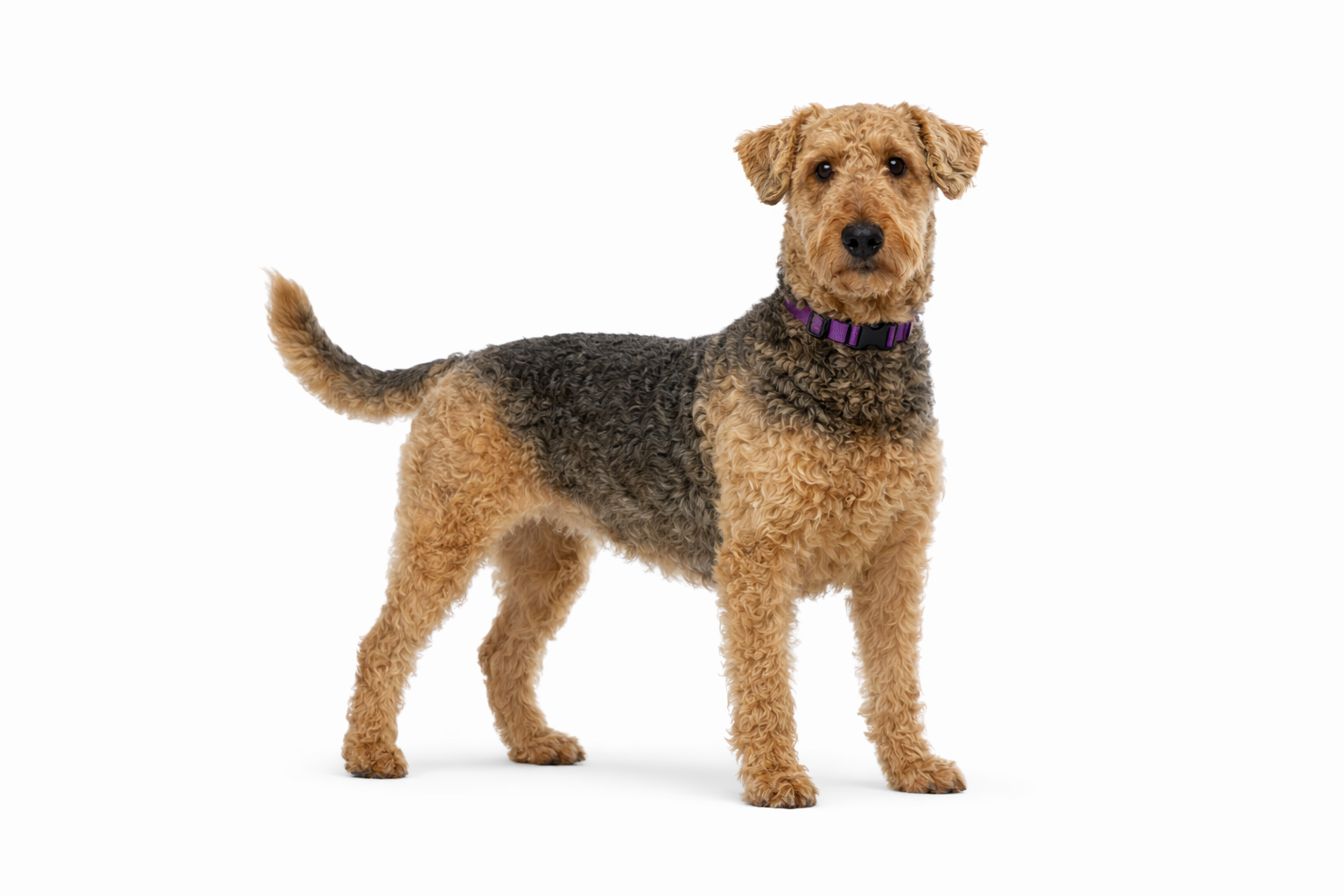}       &
 \includegraphics[width=1.5cm, trim={0, 59, 150, 59}, clip]{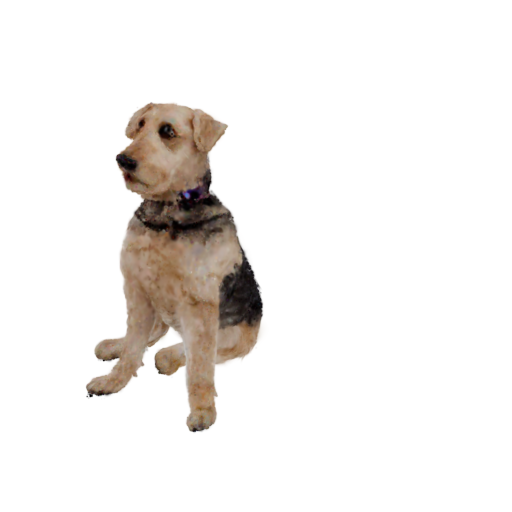} & 
\raisebox{3em}{{code}}&
  \includegraphics[width=1.5cm, trim={0, 20, 70, 40}, clip]{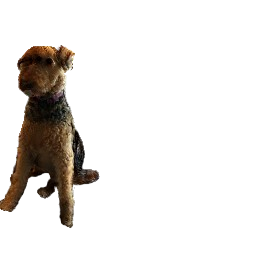} &
   \includegraphics[width=1.5cm, trim={0, 20, 70, 50}, clip]{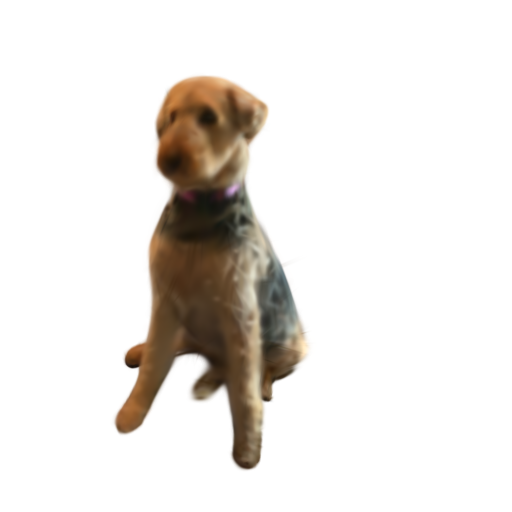} \\
\end{tabular}
}
\vspace{-0.25cm}
\caption{We present a visual evaluation of the proposed approach across different motion settings. For the GART dataset, we employ direct pose manipulation by modifying the motion parameters $\Theta$ corresponding to the \textit{walking} sequence provided by GART. For CoP3D, we apply image driven modifications, transforming the canonical pose to reference one. Our approach delivers clear visual improvements despite not being trained on the full video sequences (> 200 frames) and showing flexibility of our method in adapting to in-the-wild dog pose.}
\label{fig:coparision2}
\vspace{-0.5cm}
\end{figure*}

The visual appearance of the avatar is represented by the set of Gaussian splats
$
    \mathcal{G} = \{g_i \mid i = 1, \dots, N \}, \; g_i = (\mathbf{p}_i, \mathbf{R}_i, \mathbf{S}_i, o_i, \mathbf{c}_i),
$
where each primitive $g_i$ is defined by its position $\mathbf{p}_i\in\mathbb{R}^3$, rotation $\mathbf{R}_i\in\mathbb{R}^{3\times 3}$, scaling vector $\mathbf{S}_i$, opacity $o_i \in [0, 1]$, and color coefficients $\mathbf{c}_i$.

Reconstructing an articulated 3D animal avatar from a single image is a challenging task due to missing geometric information. To overcome this, we first use a feed-forward generative image-to-3D model, such as TripoSG~\cite{li2025triposg}, to map the single image into a 3D representation. For comparison of different models see Fig.~\ref{fig:inputdataset}.

While this generated 3D model provides a strong visual baseline, it lacks the articulation. Consequently, it is a static, non-editable geometry that can not be posed or animated. To address this limitation and enable animation, we transition to a parameterized representation, where Gaussians will be parameterized by the underlying SMAL mesh. We use the non-parametric mesh solely as a proxy to create a dense multi-view dataset. By rendering this mesh from numerous camera viewpoints $\mathbf{E}_j$ distributed uniformly on a sphere, we create a pseudo-ground truth dataset: $
    \mathcal{D}_{gt} = \{ (\mathbf{I}_j, \mathbf{E}_j) \mid j = 1, \dots, M \},
$
where $\mathbf{I}_j$ is the rendered image from viewpoint $\mathbf{E}_j$. This dataset then serves as the multi-view supervision for optimizing our SMAL+GS model, effectively learning to parameterize the geometry and appearance from this synthetic data.

\vspace{-0.5cm} 
\subsection{Joint Training of SMAL and GS}

\vspace{-0.25cm}
Our approach requires the joint optimization of the SMAL-based mesh and Gaussian Splatting representation. The model consists of two stages.

\noindent\textbf{Stage I: Bound Optimization} To ensure the 3D Gaussians capture the geometry of the animal, we initially constrain them to the SMAL mesh surface. Instead of optimizing the global 3D coordinates directly, each Gaussian's position is parameterized using learnable barycentric coordinates $\alpha$ relative to vertices $\mathbf{v}_0, \mathbf{v}_1, \mathbf{v}_2\in\mathcal{V}$ of the assigned face $\mathbf{p} = \sum_{i=0}^2 \alpha_i \mathbf{v}_i$, each face is assigned a fixed number of splats. The covariance of each splat depends only on the geometry of its assigned face. Let $\mathbf{c}$ be the face centroid. We compute the normal vector $\mathbf{e}_0$:
$
    \mathbf{n} = (\mathbf{v}_1 - \mathbf{v}_0) \times (\mathbf{v}_2 - \mathbf{v}_0), \quad \mathbf{e}_0 = \frac{\mathbf{n}}{\|\mathbf{n}\|}
$
The second vector $\mathbf{e}_1$ depends on the triangle centroid and on of its vertices:
$    \mathbf{t}_1 = \mathbf{v}_1 - \mathbf{c}, \quad \mathbf{e}_1 = \frac{\mathbf{t}_1}{\|\mathbf{t}_1\|}.
$
The vectors $\mathbf{e}_0$ and $\mathbf{e}_1$ are already orthogonal, to obtain the last basis vector $\mathbf{e}_2$ we use a single Gram-Schmidt step: 
$
    \mathbf{t}_2 = \mathbf{v}_2 - \mathbf{c}, \quad \mathbf{t}_2^\perp = \mathbf{t}_2 - \langle\mathbf{t}_2, \mathbf{e}_0\rangle\mathbf{e}_0 - \langle\mathbf{t}_2,\mathbf{e}_1\rangle\mathbf{e}_1, \quad \mathbf{e}_2 = \frac{\mathbf{t}_2^\perp}{\|\mathbf{t}_2^\perp\|}.
$
The local rotation matrix is defined as $\mathbf{R} =[\mathbf{e}_0 \mid \mathbf{e}_1 \mid \mathbf{e}_2]$, and the scale matrix 
$\mathbf{S} =[\epsilon, \frac{1}{2}\|\mathbf{t}_1\|, \frac{1}{2}\langle\mathbf{t}_2, \mathbf{e}_2\rangle]$, where $\epsilon$ is a small constant close to 0. During this stage, the Gaussian parameters are jointly optimized alongside the SMAL parameters using the pseudo ground truth dataset and a loss function $\mathcal{L}_{bound}$. This objective consists of photometric loss and structural regularization. The photometric term is a standard combination of $L_1$ distance and Structural Similarity (SSIM):
\begin{equation}
    \mathcal{L}_{rgb} = (1 - \lambda_{dssim}) |\mathbf{I} - \mathbf{I}_{gt}| + \lambda_{dssim} (1 - \text{SSIM}(\mathbf{I}, \mathbf{I}_{gt}))
\end{equation}

To maintain a well-behaved mesh topology, we apply edge-length ($\mathcal{L}_{edge}$) and Laplacian smoothing ($\mathcal{L}_{lap}$) constraints to the SMAL mesh. Additionally, to avoid degenerate skeletal configurations as the pose optimizes away from the canonical state $\hat{\theta}$, we penalize deviations of specific articulated joints $\theta_{sub}$ (e.g., legs, mouth), while an $L_2$ norm restricts the vertex offsets $\mathbf{d}$. 

\begin{wraptable}{r}{0.6\textwidth}
\vspace{-1cm} 
\caption{Comparison of GART and our method using CLIP-Score between renders of \textit{walking} animation and reference image of the dog.  \our{} shows greater similarity to the input image}
 \centering
\setlength{\tabcolsep}{4.3pt}
{\fontsize{6.8pt}{11pt}\selectfont{
        \begin{tabular}{lcc|cc} 
& \multicolumn{2}{c|}{GART dataset} & \multicolumn{2}{c}{CoP3D} \\  \cmidrule(lr){2-3}  \cmidrule(lr){4-5}
methods & \makecell{Alaskan \\ Malamute} & Corgi & Husky &  \makecell{Airedale \\ Terrier} \\ \hline
GART &  0.578 & 0.644 & 0.820 & 0.674 \\
\textbf{\our{} }& \textbf{0.598} & \textbf{0.738} & \textbf{0.850} & \textbf{0.752} \\
\end{tabular}
}\label{tab:comparisonwalking}}
\vspace{-0.7cm}
\end{wraptable}

\noindent The final component of this early stage is the opacity term:
\vspace{-0.3cm}
\begin{equation}
    \mathcal{L}_{opac} = \frac{1}{N} \sum_{i=1}^N -o_i.
\end{equation}

By minimizing the negative opacity, we force the surface-bound Gaussians to become as opaque as possible ($o_i \to 1$), preventing semi-transparent artifacts during the shape alignment stage weighted by $\lambda_{opac}=0.001$. The total loss for Stage I is thus defined as:
\begin{equation}
    \mathcal{L}_{bound} = \mathcal{L}_{rgb} + \mathcal{L}_{edge} + \mathcal{L}_{lap} + \|\mathbf{d}\|_2^2 + \|\theta_{sub} - \hat{\theta}_{sub}\|_2^2 + \lambda_{opac}\mathcal{L}_{opac}
\end{equation}

\begin{figure}[t]
\centering
\setlength{\tabcolsep}{1pt}
{\fontsize{6.8pt}{11pt}\selectfont
\begin{NiceTabular}{cccccc}
  Input & {Sitting}  & \hspace{4mm} & 
    {Walking}   & \hspace{4mm} & 
  {Give a paw}  \\
\Block{3-1}{
\includegraphics[width=0.7cm]{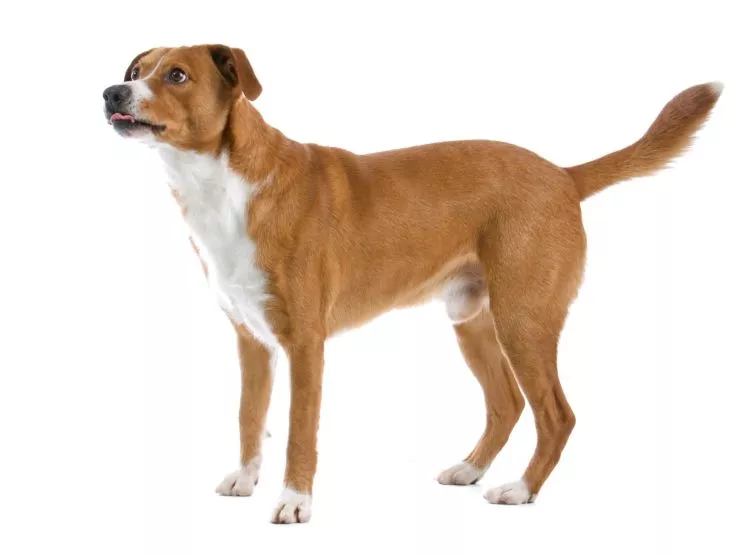}
} & 
\includegraphics[width=0.7cm]{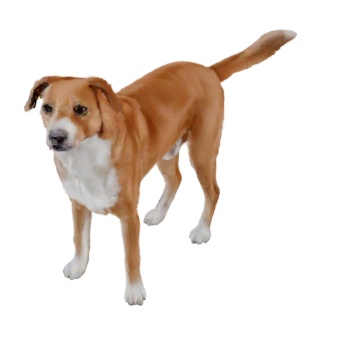} 
\includegraphics[width=0.7cm]{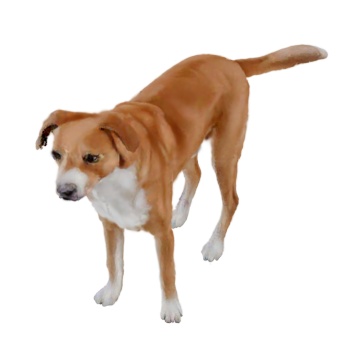} 
\includegraphics[width=0.7cm]{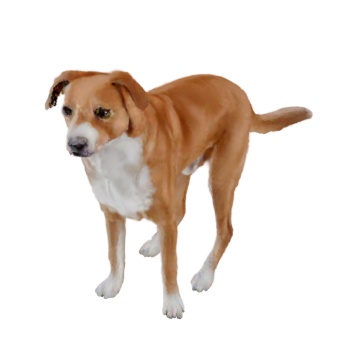} 
\includegraphics[width=0.7cm]{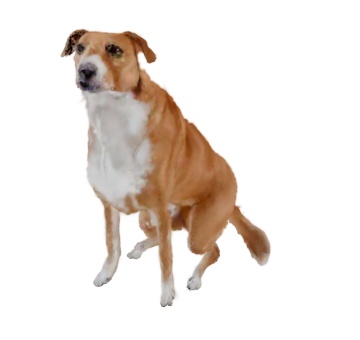}  & &
\includegraphics[width=0.7cm]{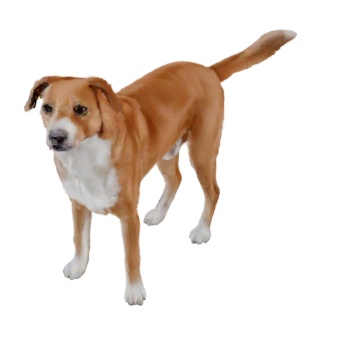} 
\includegraphics[width=0.7cm]{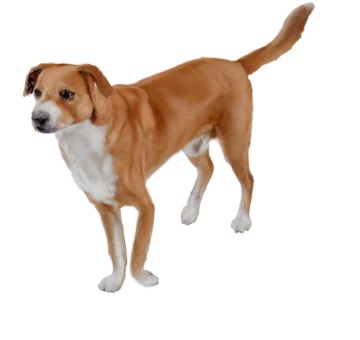} 
\includegraphics[width=0.7cm]{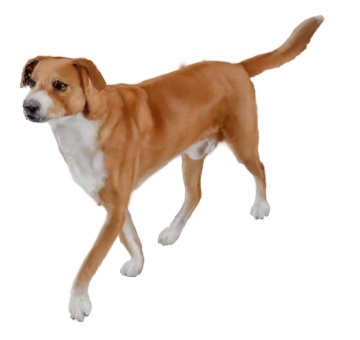} 
\includegraphics[width=0.7cm]{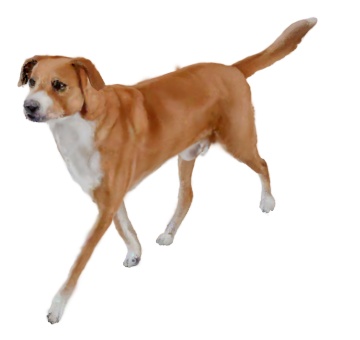}  & &
\includegraphics[width=0.7cm]{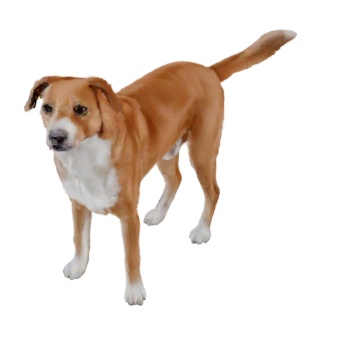} 
\includegraphics[width=0.7cm]{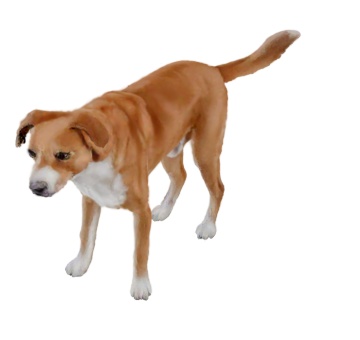} 
\includegraphics[width=0.7cm]{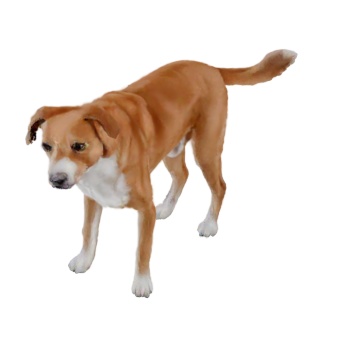} 
\includegraphics[width=0.7cm]{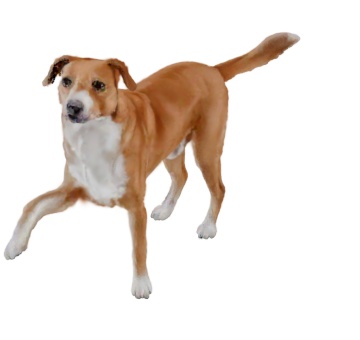}   \\
&
\includegraphics[width=0.7cm]{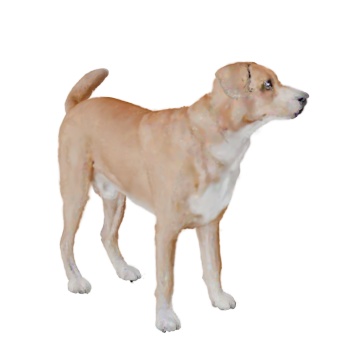} 
\includegraphics[width=0.7cm]{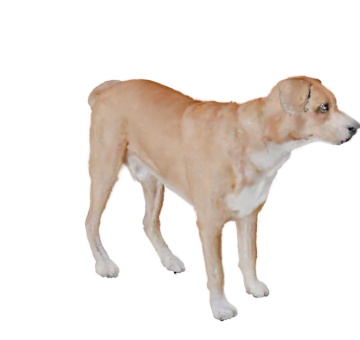} 
\includegraphics[width=0.7cm]{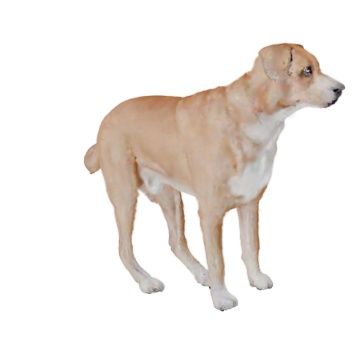} 
\includegraphics[width=0.7cm]{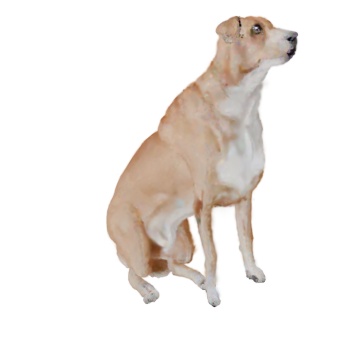}  & &
\includegraphics[width=0.7cm]{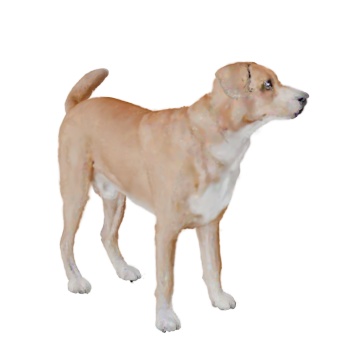} 
\includegraphics[width=0.7cm]{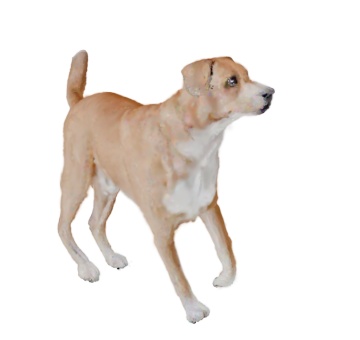} 
\includegraphics[width=0.7cm]{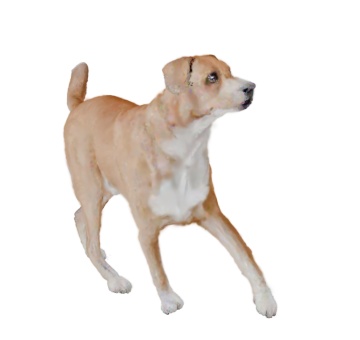} 
\includegraphics[width=0.7cm]{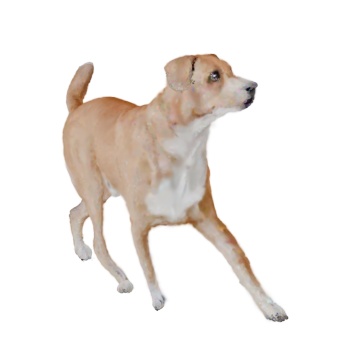}& &
\includegraphics[width=0.7cm]{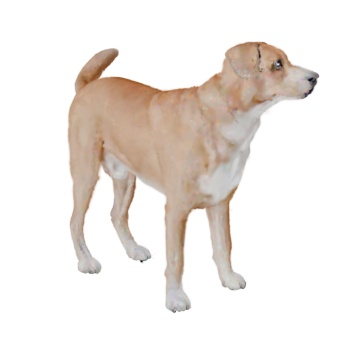} 
\includegraphics[width=0.7cm]{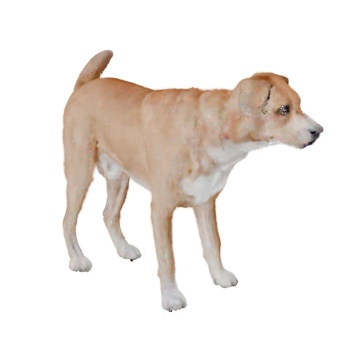} 
\includegraphics[width=0.7cm]{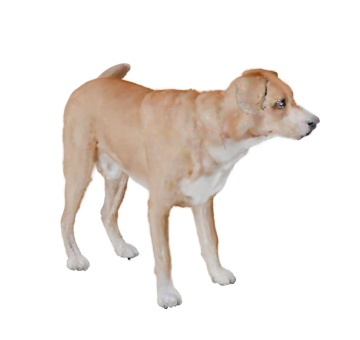} 
\includegraphics[width=0.7cm]{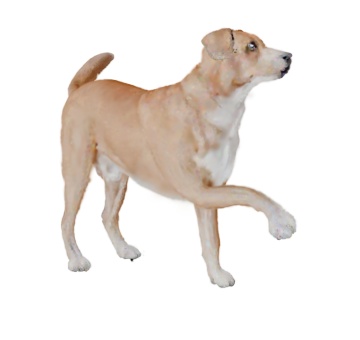}   \\
&
\includegraphics[width=0.7cm]{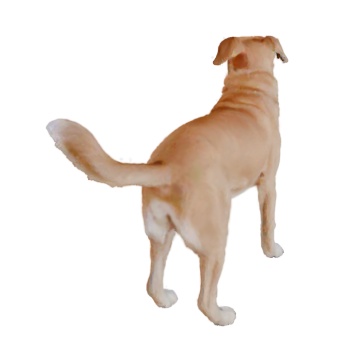} 
\includegraphics[width=0.7cm]{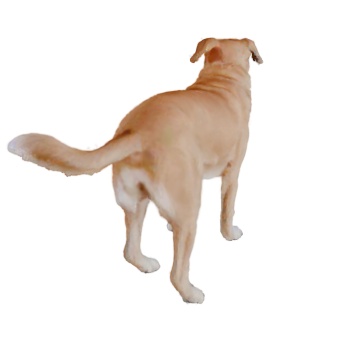} 
\includegraphics[width=0.7cm]{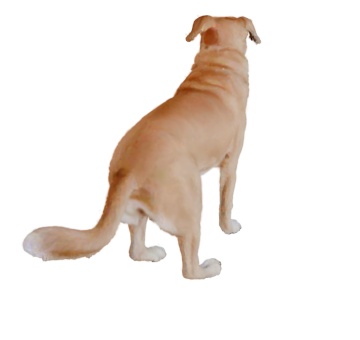} 
\includegraphics[width=0.7cm]{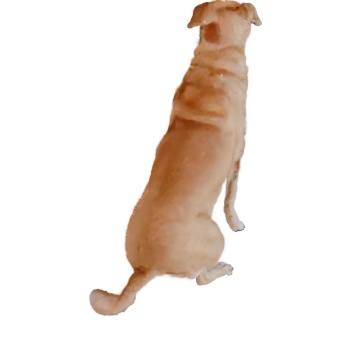}   & &
\includegraphics[width=0.7cm]{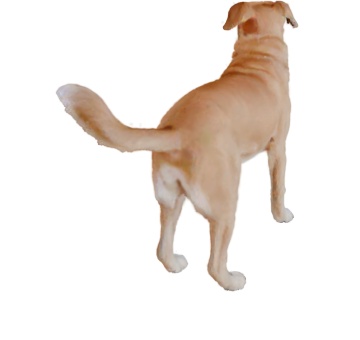} 
\includegraphics[width=0.7cm]{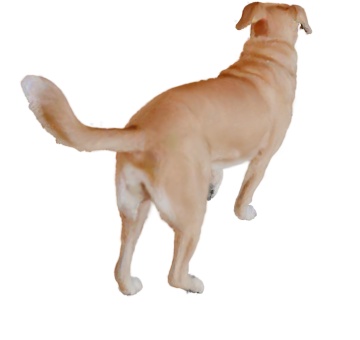} 
\includegraphics[width=0.7cm]{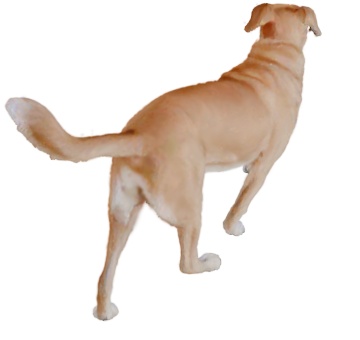} 
\includegraphics[width=0.7cm]{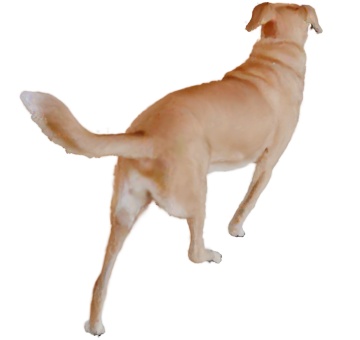}  & &
\includegraphics[width=0.7cm]{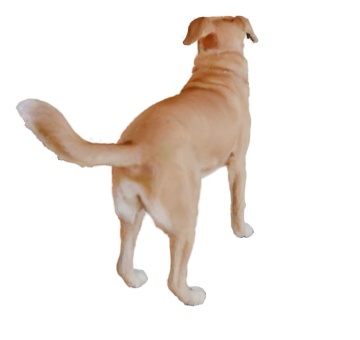} 
\includegraphics[width=0.7cm]{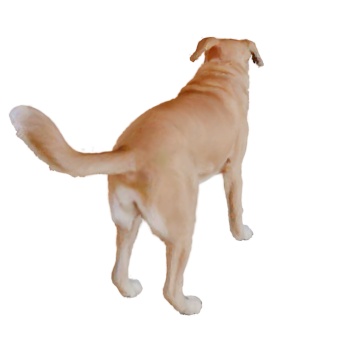} 
\includegraphics[width=0.7cm]{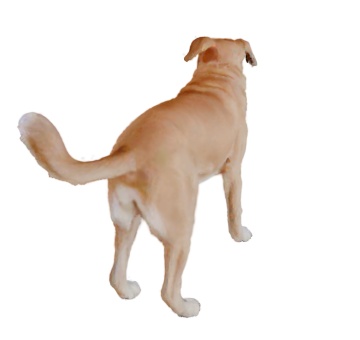} 
\includegraphics[width=0.7cm]{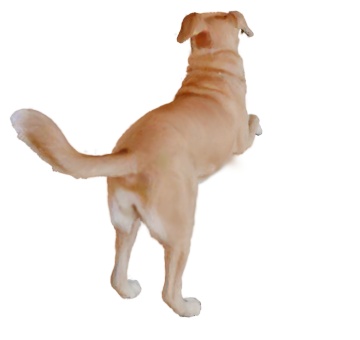}   \\
\end{NiceTabular}}
\caption{Our approach enables avatar animation for arbitrary movements using Image Driven Modifications. We first select an initial frame rendered from the chosen camera and generate a target frame via a generative model (e.g., “Change the pose of this dog; the dog is sitting”). Then FramePack is used to interpolate between these frames, creating a video of the animal movement to the target pose. Finally, ActionMesh maps the video back onto our 3D representation, using the fitted SMAL model as a prior to ensure anatomically consistent motion.}
\label{fig:animations}
\vspace{-0.5cm}
\end{figure}

\noindent\textbf{Stage II: Unbound Optimization}
While the surface-bound representation aligns the SMAL model to the general body shape, the ability to reconstruct high-detail features, such as fur, is limited when Gaussian primitives are bound to the mesh surface, see Tab.~\ref{tab:stage1dge}. Therefore, after 15000 iterations, we decouple gaussians from the mesh. The previously computed $\mathbf{p}$, $\mathbf{R}$, and $\mathbf{S}$ are detached from the barycentric formulation used in Stage I, and are optimized directly. Standard 3DGS adaptive density control is subsequently activated to reconstruct fine volumetric details. 

Since the gaussians are no longer tied to the mesh optimizing them without additional constraints could lead to the loss of geometry understanding. Consequently, to maintain structural integrity, the photometric loss ($\mathcal{L}_{rgb}$) and regularization terms ($\mathcal{L}_{edge},\,\mathcal{L}_{lap}$), as well as offset and pose penalties) remain. However, to ensure that the unbound Gaussians stay close to the underlying mesh geometry, we add a point-to-mesh distance term ($\mathcal{L}_{dist}$) that penalizes the Euclidean distance between the Gaussian centers and the closest SMAL mesh faces. Crucially, for this term we propagate gradients only to the parameters of SMAL, not positions of gaussians. This is done to let the mesh make small adjustments to better fit the animal's shape without limiting the ability of primitives to reconstruct the lacking details. Furthermore, we impose additional constraints on the parameters of Gaussian primitives. To discourage creation of large elliptical splats, we add scale regularization:

\vspace{-0.5cm}
\begin{align}
    \mathcal{L}_{s} &=\frac{1}{N} \sum_{i=1}^N ( \|\mathbf{S}_i\|_2^2 \nonumber \\ 
    + &\max(\mathbf{S}_i) - \min(\mathbf{S}_i) ).
\end{align}
\vspace{-0.8cm}

The main goal of this part is to enable smoother rotation when applying animations, which might be caused by large spiky Gaussians. In Stage II we also redefine the opacity term, which now is replaced by an entropy based regularization term: 

\begin{equation}
    \mathcal{L}_{ent} = \frac{1}{N} \sum_{i=1}^N -o_i \log(o_i).
\end{equation}


This causes the Gaussians to become fully opaque ($o_i \to 1$) or transparent ($o_i \to 0$). The comprehensive loss objective for the unbound refinement stage is thus defined as:
\begin{align*}
    \mathcal{L}_{unbound} = & \mathcal{L}_{rgb} + \mathcal{L}_{edge} + \mathcal{L}_{lap} + \|\mathbf{d}\|_2^2 + \|
    \theta_{sub} - \hat{\theta}_{sub}\|_2^2 \\ & + \lambda_{dist}\mathcal{L}_{dist} + \mathcal{L}_{s} + \lambda_{opac}\mathcal{L}_{ent},
    \addtocounter{equation}{1}\tag{\theequation}
\end{align*}
where $\lambda_{dist}=10,\;\lambda_{opac}=0.001$ in our experiments. After this stage finishes we obtain the base version of our model.
\\
\\

\begin{wrapfigure}{r}{0.6\textwidth}
\vspace{-1cm}
 \centering
\setlength{\tabcolsep}{4.3pt}
{\fontsize{6.8pt}{11pt}\selectfont
\begin{tabular}{cccccc}
\multicolumn{6}{c}{puppy + example movement} \\
 \includegraphics[width=0.07\textwidth, trim={50, 100, 50, 150}, clip]{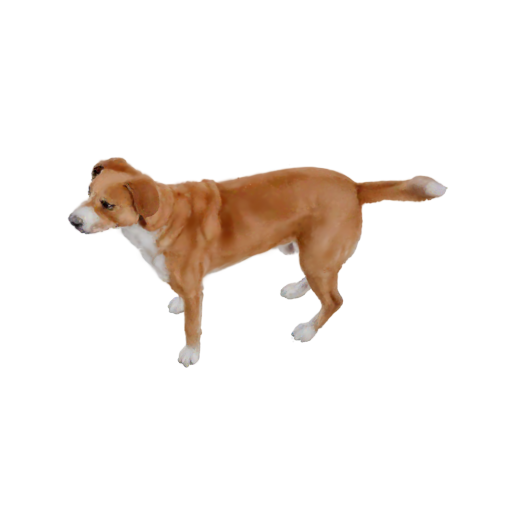} &
 \includegraphics[width=0.07\textwidth, trim={50, 100, 50, 150}, clip]{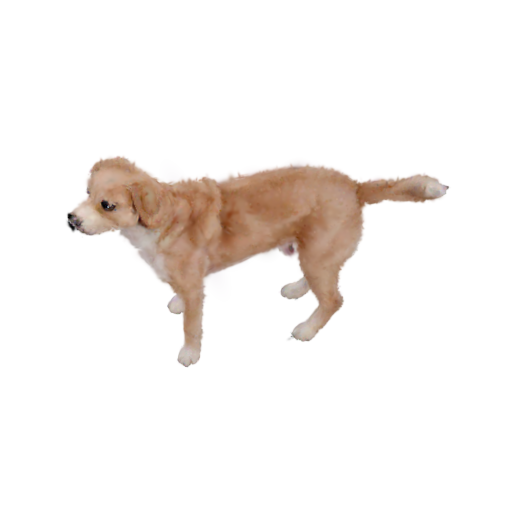} & 
 \includegraphics[width=0.07\textwidth, trim={50, 50, 50, 50}, clip]{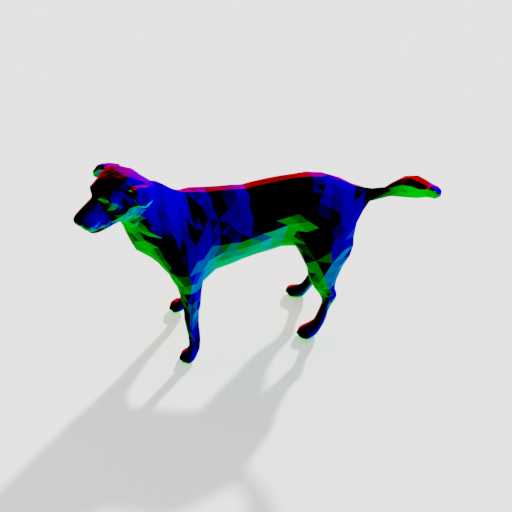} &
 \includegraphics[width=0.07\textwidth, trim={50, 100, 50, 100}, clip]{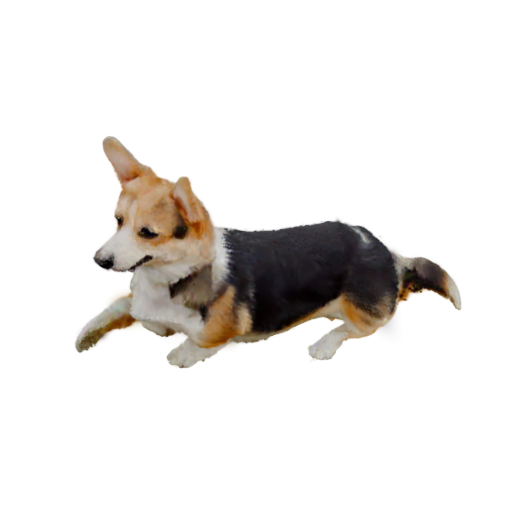} &
 \includegraphics[width=0.07\textwidth, trim={50, 100, 50, 100}, clip]{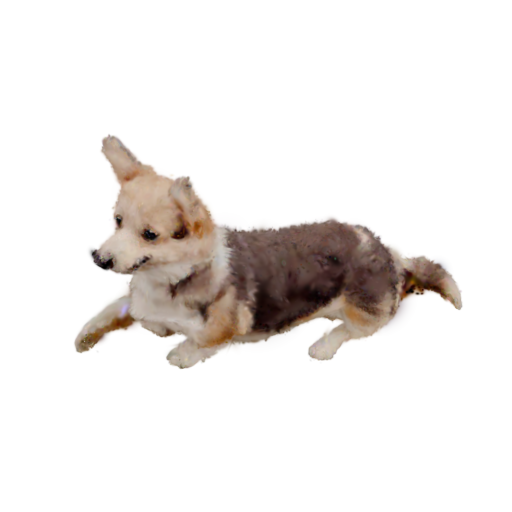} &    \includegraphics[width=0.07\textwidth, trim={50, 50, 50, 50}, clip]{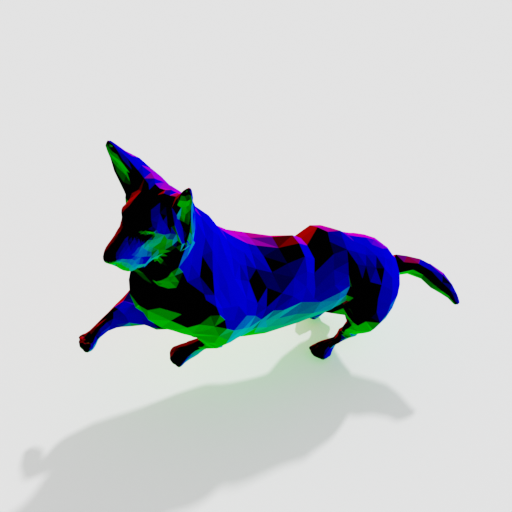} \\
    \includegraphics[width=0.07\textwidth, trim={50, 100, 50, 150}, clip]{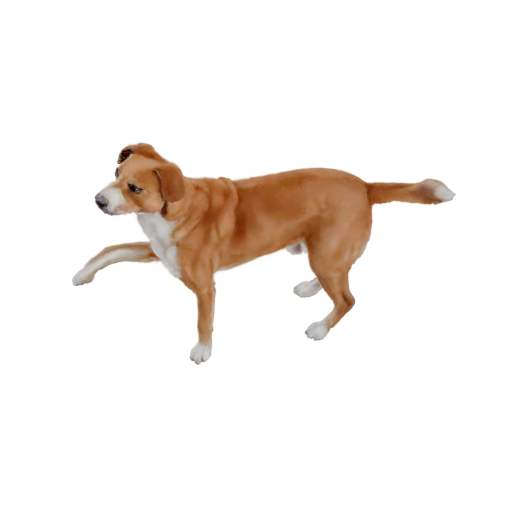} &
 \includegraphics[width=0.07\textwidth, trim={50, 100, 50, 150}, clip]{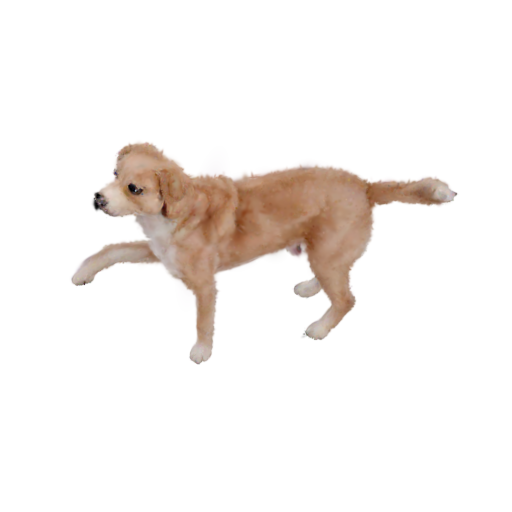}  & 
   \includegraphics[width=0.07\textwidth, trim={50, 50, 50, 50}, clip]{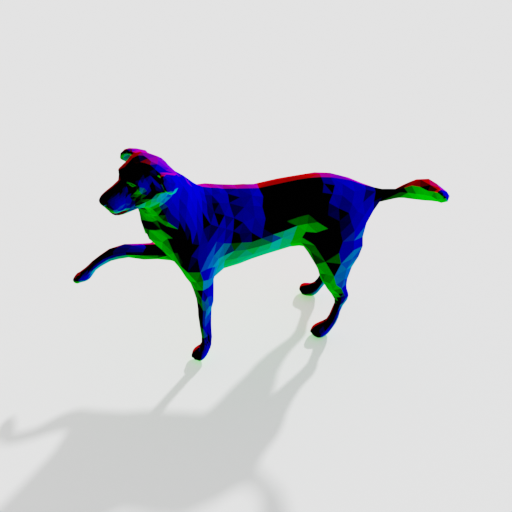}  &
 \includegraphics[width=0.07\textwidth, trim={50, 100, 50, 100}, clip]{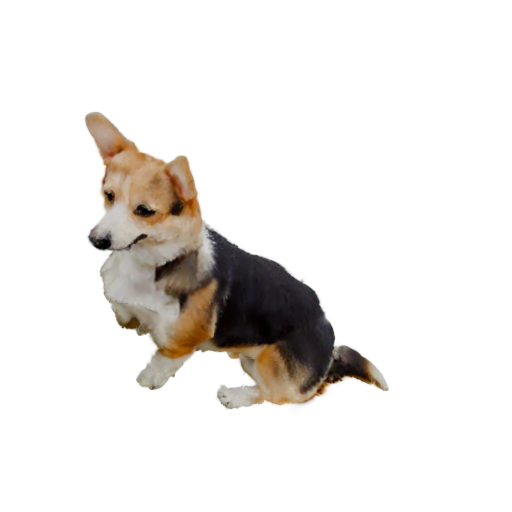} &
 \includegraphics[width=0.07\textwidth, trim={50, 100, 50, 100}, clip]{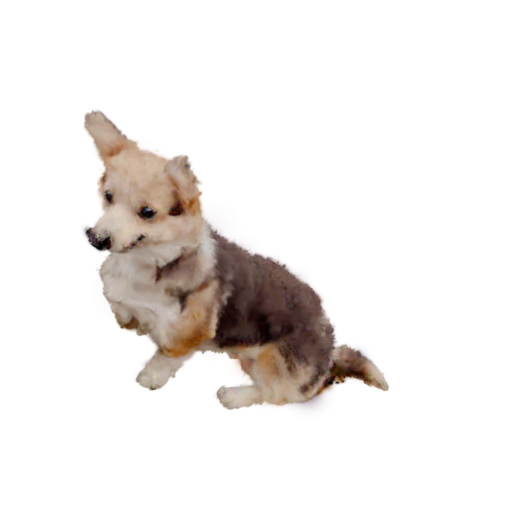}  &
  \includegraphics[width=0.07\textwidth, trim={50, 50, 50, 50}, clip]{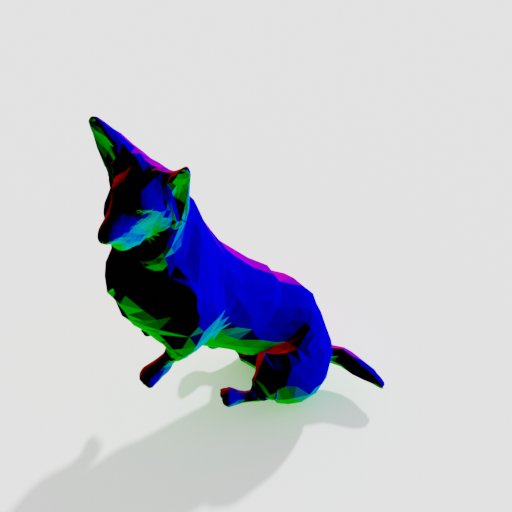} 
\\
 \multicolumn{3}{c}{marble + sit}  & \multicolumn{3}{c}{manga + sit } \\
 \includegraphics[width=0.07\textwidth, trim={50, 90, 50, 100}, clip]{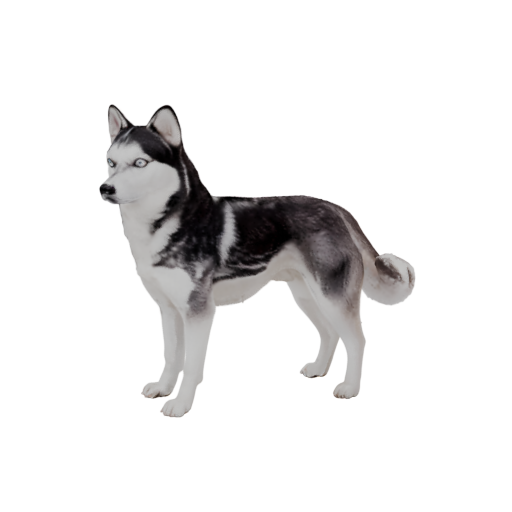}  &
 \includegraphics[width=0.07\textwidth, trim={50, 90, 50, 100}, clip]{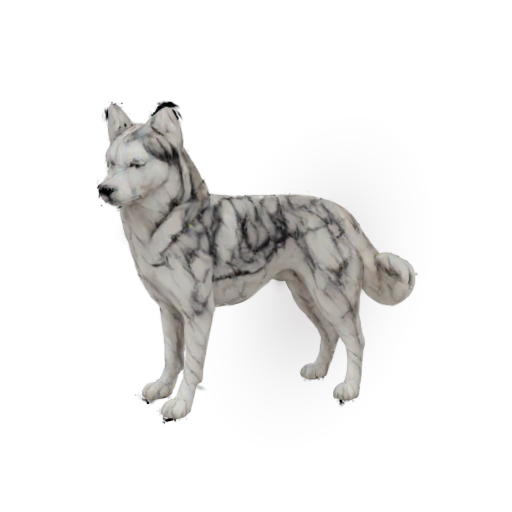} & 
\includegraphics[width=0.07\textwidth, trim={50, 50, 50, 50}, clip]{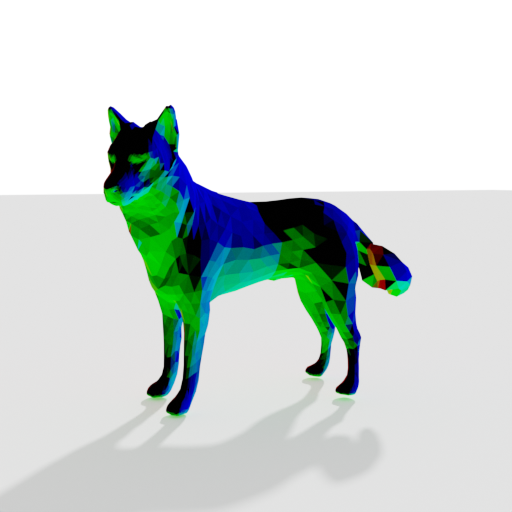}  &
 \includegraphics[width=0.07\textwidth, trim={50, 50, 50, 50}, clip]{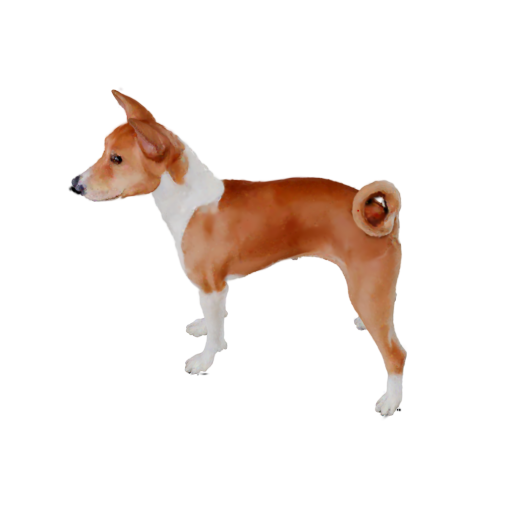} &
 \includegraphics[width=0.07\textwidth, trim={50, 50, 50, 50}, clip]{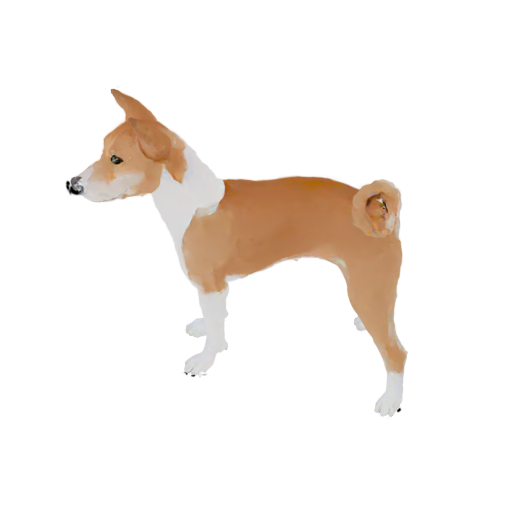} &   \includegraphics[width=0.07\textwidth, trim={50, 50, 50, 50}, clip]{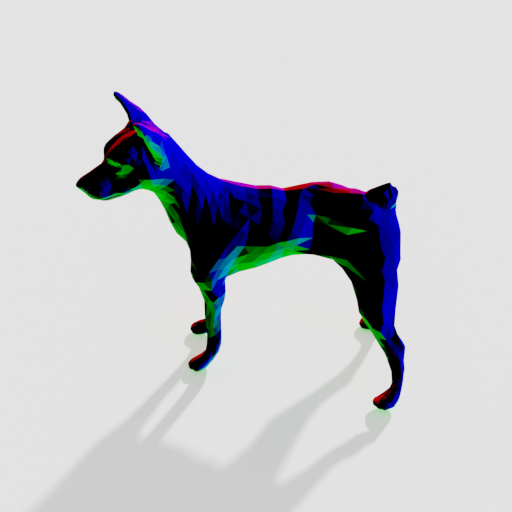} \\
 \includegraphics[width=0.07\textwidth, trim={100, 100, 100, 100}, clip]{imgs/styles/dog10/deafult_view_86_00015.png} &
 \includegraphics[width=0.07\textwidth, trim={100, 100, 100, 100}, clip]{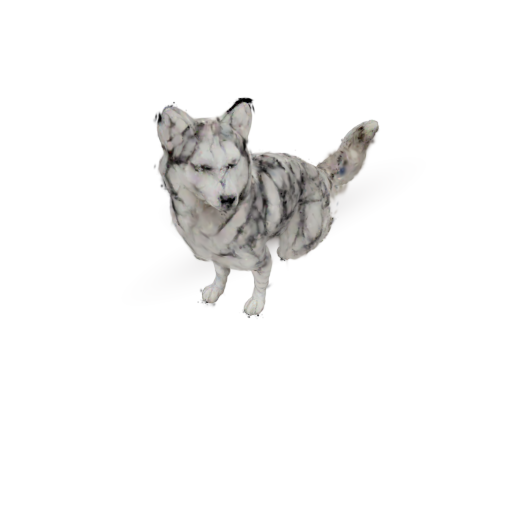} & 
\includegraphics[width=0.07\textwidth, trim={50, 100, 100, 50}, clip]{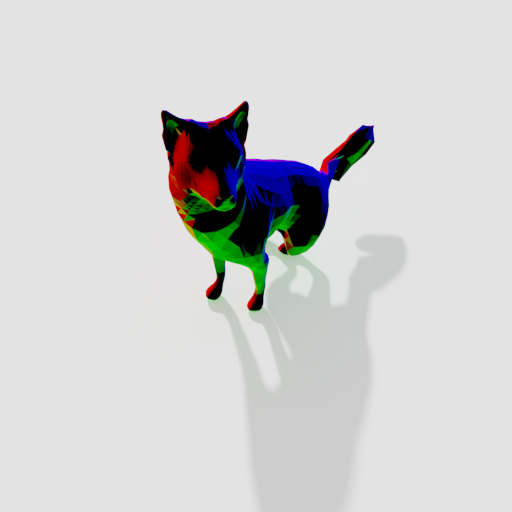} &
 \includegraphics[width=0.07\textwidth, trim={50, 50, 50, 50}, clip]{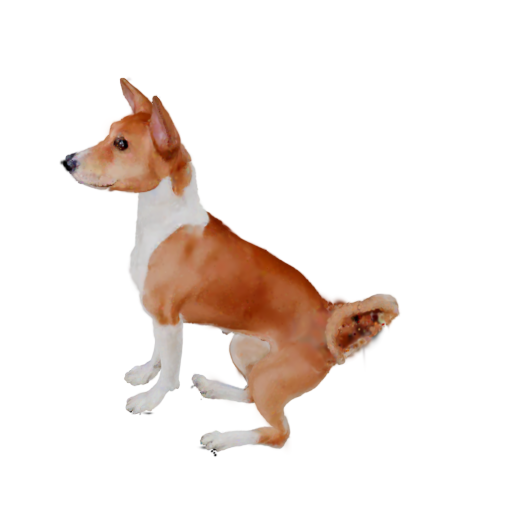} &
 \includegraphics[width=0.07\textwidth, trim={50, 50, 50, 50}, clip]{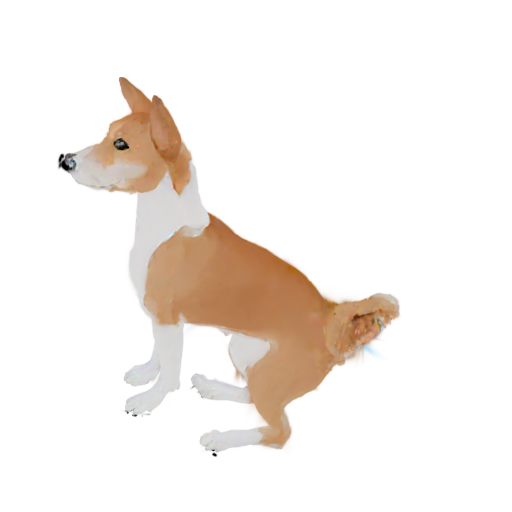}  &  \includegraphics[width=0.07\textwidth, trim={50, 50, 50, 50}, clip]{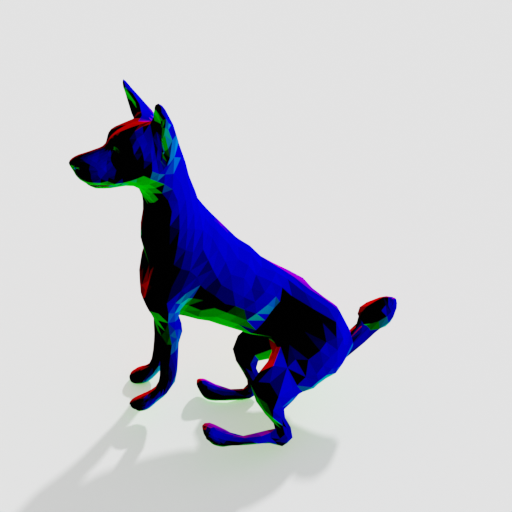}
\end{tabular}
}
\caption{Our pipeline supports both visual changes and pose animation based on fitted SMAL mesh. By attaching Gaussians to the SMAL mesh our method allows flexible avatar animation across different poses and appearances.}
\label{fig:styleandmove}
\end{wrapfigure}
\noindent\textbf{Face-Based Skinning for Animation}
Following the completion of the two-stage optimization, the unbound Gaussians, each with parameters: position $\mathbf{p}$, rotation matrix $\mathbf{R}$, and scale $\mathbf{S}$, must be re-associated with the articulated SMAL proxy mesh to enable novel pose animations. We achieve this by baking skinning weights into the Gaussians and embedding them within local coordinate systems defined by the mesh faces. 

For a given reference mesh face $k$ with vertices $\mathbf{v}_0, \mathbf{v}_1, \mathbf{v}_2$, we define an orthonormal local face basis:
\begin{equation}
    \mathbf{x}_k = \frac{\mathbf{v}_1 - \mathbf{v}_0}{\|\mathbf{v}_1 - \mathbf{v}_0\|}, \quad 
    \mathbf{z}_k = \frac{(\mathbf{v}_1 - \mathbf{v}_0) \times (\mathbf{v}_2 - \mathbf{v}_0)}{\|(\mathbf{v}_1 - \mathbf{v}_0) \times (\mathbf{v}_2 - \mathbf{v}_0)\|}, \quad 
    \mathbf{y}_k = \mathbf{z}_k \times \mathbf{x}_k
\end{equation}
Crucially, this basis defines a correct rotation matrix given as: $
     \mathbf{R}_k = \begin{bmatrix} \mathbf{x}_k & \mathbf{y}_k & \mathbf{z}_k \end{bmatrix}
$
Additionally, we record the reference edge length sum of the face $L_{ref, k}$. 

For each optimized Gaussian with final global position $\mathbf{p}$ and rotation $\mathbf{R}$, we find the $K=10$ nearest SMAL mesh faces. We compute skinning weights $w_k$ based on the inverse distance $d_k$ to these face centers:
$
    w_k = \frac{d_k^{-1}}{\sum_{j=1}^K d_j^{-1}}
$
The Gaussian's parameters are then transformed and baked into the local coordinate frame of each assigned face $k$ with centroid $\mathbf{c}_k$:
\begin{equation}
    \mathbf{p}_{loc, k} = \mathbf{R}_k^\top (\mathbf{p} - \mathbf{c}_k), \quad \mathbf{R}_{loc, k} = \mathbf{R}_k^\top \mathbf{R}
\end{equation}

\noindent\textbf{Skinning Animation} 
During inference, modifying the SMAL pose $\theta$ updates the mesh geometry, resulting in new face centroids $\mathbf{c}_k^{new}$, local basis rotations $\mathbf{R}_k^{new}$, and edge length sum $L_k^{new}$. The Gaussians are then deformed using an analogue to Linear Blend Skinning (LBS). The new position $\mathbf{p}_{new}$ is calculated as a linear combination of the local coordinates projected through the updated mesh faces:
\begin{equation}
    \mathbf{p}_{new} = \sum_{k=1}^K w_k \left( \mathbf{c}_k^{new} + \mathbf{R}_k^{new} \mathbf{p}_{loc, k} \right)
\end{equation}

Similarly, each binding face defines a new rotation matrix $\mathbf{R}_{glob, k} = \mathbf{R}_k^{new} \mathbf{R}_{loc, k}$. Let $\mathbf{q}_{glob, k}$ denote the quaternion representation of this matrix. The final animated rotation $\mathbf{q}_{new}$ of the corresponding Gaussian is obtained by blending these quaternions:
\begin{equation}
    \mathbf{q}_{new} = \text{normalize}\left( \sum_{k=1}^K w_k \mathbf{q}_{glob, k} \right)
\end{equation}

\begin{wraptable}{r}{0.55\textwidth}
\vspace{-0.35cm}
\caption{\our{} consists of two main stages. We demonstrate that stage 2 is essential for improving the quality of synthetic dataset reconstruction. Furthermore, we show that incorporating an improved appearance stage as Final has a positive impact on the CLIP score.}
 \centering
\setlength{\tabcolsep}{4.3pt}
{\fontsize{6.8pt}{11pt}\selectfont{
        \begin{tabular}{lcc|cc} 
& \multicolumn{2}{c|}{PSNR} & \multicolumn{2}{c}{CLIP-Score} \\  \cmidrule(lr){2-3}  \cmidrule(lr){4-5}
         after & stage 1 & stage 2 & stage 2 & \makecell{Final} \\ \hline
        \includegraphics[width=0.07\textwidth, trim={0, 40, 0, 40}, clip]{imgs/inputs/dog2.png}&  29.27 &  \textbf{42.97 }&  0.85 &  \textbf{0.88 }\\
        \includegraphics[width=0.07\textwidth, trim={0, 0, 0, 20}, clip]{imgs/inputs/dog6.jpg} &  27.25 &  \textbf{41.81 }&  0.86 & \textbf{ 0.89} \\
        \includegraphics[width=0.07\textwidth, trim={100, 100, 100, 100}, clip]{imgs/inputs/dog11.png} &  30.52 &  \textbf{43.73 }&  0.86 &  \textbf{0.88 }\\
\end{tabular}
}\label{tab:stage1dge}}
\vspace{-1.5cm}
\end{wraptable}

Finally, to account for stretching of the underlying mesh during movement, the physical scale of each Gaussian is adjusted proportionally to the relative change in the edge length sum of its corresponding faces. The new scale vector $\mathbf{S}_{new}$ is computed as:
\begin{align*}
    s_{new} &= \sum_{k=1}^K w_k \sqrt{\frac{L_{new, k}}{L_{ref, k}}} \\
    \mathbf{S}_{new} &= s_{new}\,\mathbf{S}  
\end{align*}
This skinning formulation translates the low-dimensional skeletal animation of the underlying SMAL mesh into a continuous deformation of the 3D Gaussian field. The remaining parameters of each Gaussian remain unchanged.

\vspace{-0.25cm}
\section{Application: Editing and Animation of \our{}} 

Since our model combines a SMAL with 3D Gaussians, the resulting animal avatar is highly editable. Our representation avoids the limitations of static 3D geometry by providing a controllable framework that can be modified in two primary ways: through direct pose manipulation and image driven modifications.

\noindent\textbf{Direct Pose Manipulation}
The SMAL model allows for intuitive edits. Because the Gaussian splats are rigged to the underlying mesh faces, they maintain spatial consistency during any deformation. By manipulating the the parameters~$\Theta$, consequently updating the mesh structure, the Gaussians also update.

Specifically, by modifying the pose $\theta$, the model can be articulated into any skeletal configuration. This enables the motion transfer, where skeletal animations can be transferred between avatars. Since all our avatars share the same SMAL topology, we can apply motion sequences directly to any reconstructed breed. We evaluate the effectiveness of this transfer by comparing our animated results with animations based on GART. These comparisons demonstrate that our method preserves visual appearance better than existing benchmarks.

\noindent\textbf{Image Driven Modifications} 
The second way to animate the avatar is through image driven. This allows users to create new animations by defining a target state for the animal. This state can be obtained by either providing an existing reference image of a dog in desired position, or using a generative text-to-image model using textual description of the modification.

This final state has to be then translated into motion. This is achieved by utilizing FramePack~\cite{zhang2025frame}, specifically we used the HunyuanVideo~\cite{kong2024hunyuanvideo} transformer based version, to generate a time-consistent video between two images. The initial input image serves as the starting point, and the chosen target image plays the role of the last frame, to help the generation process we also provide a text prompt \textit{``The creature is moving.''}. FramePack interpolates between these two frames, and generates a video of the animal moving into the target pose.

Finally, we use ActionMesh~\cite{sabathier2026actionmesh} to map the video back to our 3D representation. ActionMesh takes the generated video and SMAL mesh learned by our method as inputs, and finds the deformed mesh for subsequent frames of the video starting with its input. Because these resulting meshes maintain the same topology as our original SMAL proxy, the 3D Gaussians can follow the deformation sequence. This enables our pipeline to perform complex animations, based on visual guidance. 

\noindent\textbf{Improved appearance} To further enhance the visual quality of the reconstructed avatar, we employ the Direct Gaussian Editor (DGE)~\cite{chen2024dge} as a final refinement step. The generated dataset based solely on the initial image-to-3D model, might lack high-frequency details of a real animal's appearance. To address this, we apply DGE guided by the text prompt ``\textit{High quality dog.}'' By leveraging its multi-view consistent editing capabilities, we inject strong generative priors directly into the optimized 3D Gaussians. This step effectively improves details, textures, and the overall photorealism of the resulting animatable avatar.

Additionally, as an application, DGE can be used again to the modified already corrected Gaussians. Using a new prompt, it alters their visual attributes, producing a unique appearance while preserving the underlying structure. We refer to the DGE-based improved stage as \textit{Final}, reflecting its role as the last correction step in the pipeline.

\vspace{-0.25cm}
\section{Experiments}
\vspace{-0.25cm}

In this section, we evaluate our proposed framework for single image to 3D animal reconstruction and animation. We demonstrate the effectiveness of our method through qualitative comparisons with state-of-the-art baselines, show the performance of our face-based skinning algorithm for complex animations, analyze the impact of different image-to-3D generation backbones, and highlight the possibility of stylized editing without hindering the editing ability\footnote{\url{https://github.com/piotr310100/SMAL-pets}}.

\noindent\textbf{Experimental Setup}
Our pipeline is implemented in PyTorch using the standard 3D Gaussian Splatting rasterizer. For the default image-to-3D generation step, we utilize TripoSG~\cite{li2025triposg} to create the initial multi-view dataset. During training Stage I (Bound Optimization) runs for 15000 iterations to find the SMAL skeletal alignment, followed by Stage II (Unbound Optimization) which runs for an additional 25000 iterations, with densification for the first 15000 iterations, to improve reconstruction of details.

\noindent\textbf{Impact of Image-to-3D Backbones}
Because our pipeline is agnostic to the specific generative model used to obtain initial 3D object from a single image, we experiment with three different state-of-the-art image-to-3D backbones: TripoSG, Trellis, and SAM3D. As visualized in Fig.~\ref{fig:inputdataset}, the choice of backbone significantly influences the level of detail and overall topological structure. SAM3D and TRELLIS can occasionally produce spiky artifacts visible in the obtained basenji dog. Moreover, they struggle with elements such as eyes, visible in the example of a husky. Consequently, they slightly lose the similarity to the input image. TripoSG on the other hand, offers consistent textures, while also preserving geometric structure without spiky artifacts. Crucially, all of the mentioned methods can lack high-frequency details, such as fur. As visible in the highlighted rows, this effect can be reduced by our method with the use of additional post processing. Additionally, we calculating the CLIP score between the input image and the out \our{} renders, averaged across all test views after stage 2 and Final using the appropriate backbone, see Fig.~\ref{fig:inputdataset} and Tab~\ref{tab:backbones}. TripoSG shows a better fit for our pipeline.


\noindent\textbf{Dataset single images:} To the best of our knowledge DogRecon is the only method that reconstructs dog avatars from a single image. However, as the official code was not publicly available at the time of submission, we are evaluating based on the same dog cases described in their paper to ensure the most accurate comparison possible. Additionally, we evaluate \our{} on the GART and Cop3D~\cite{sinha2023cop3d} datasets, which consist of long monocular videos depicting dogs. The GART dataset presents more dynamic videos. In our method, we select representative frames where the object is clearly visible to initiate reconstruction. 

\noindent\textbf{Single image-to-Avatar }
We present a qualitative comparison with DogRecon in terms of reconstructing a dog avatar from a single image. see Fig.~\ref{fig:comparisionwithdocracon}. Both methods present the reconstructed avatar using three-dimensional Gaussian primitives and a parametric SMAL model, allowing the avatar to be animated naturally by directly manipulating the parameters $\Theta$. We evaluate generalization to a novel pose by modifying the motion parameters $\Theta$ corresponding to the \textit{walking} sequence provided by GART. Our method captures the details of the dog with greater accuracy, especially in the context of fur, see Fig.~\ref{fig:dogreconvsour}.

\noindent\textbf{Comparisons with SOTA} A comparison with SOTA methods is shown in Fig.~\ref{fig:coparision2}. In the case of the GART dataset. The DogRecon results are taken directly from the original publication. To enable a direct comparison, we performed pose manipulation by modifying the motion parameters $\Theta$ corresponding to the \textit{walking} sequence provided by GART. Tab.~\ref{tab:comparisonwalking} shows CLIP-Score between renders of \textit{walking} animation and reference image of the dog (see first column in Fig.~\ref{fig:coparision2}).  A numerical comparison with Dogrecon is not possible due to the lack of code. 

On Cop3D for GART and AnimalAvatar (AniAv.), the comparable view does not represent a new dog pose, as both methods are trained on full video material. To ensure comparability, we apply text modifications from the training view to the selected frame, matching the reconstructed result to the target pose. We demonstrate that our method allows the dog avatar to be adjusted to any selected position, in particular realistic ones. 

\noindent\textbf{Applications}
We select representative real-world motions that a dog can naturally perform, including \textit{sitting}, \textit{walking}, and \textit{giving a paw}. Fig.~\ref{fig:animations} illustrates the application of our pipeline to synthesize a novel. The Image-Driven Modifications procedure is employed to generate realistic motion variations tailored to the given avatar. We demonstrate that the modifications are consistent across different camera views and provide high visual quality. A more detailed analysis is provided in the supplementary material.

Since Gaussians surround the underlying SMAL mesh and each of them can be represented in the local coordinate system of the corresponding faces, their properties and number can be freely changed. This is demonstrated in Fig.~\ref{fig:teaser} and Fig.~\ref{fig:styleandmove}, where their modification does not hinder the editing of the appearance or the application of animation. This is a result of the pre-fitted SMAL mesh being always the same for a given avatar, and new local coordinate systems can be recalculated for new Gaussians.

\begin{figure}[t]
 \centering
  \includegraphics[width=\textwidth, trim={0, 0, 0, 0}, clip]{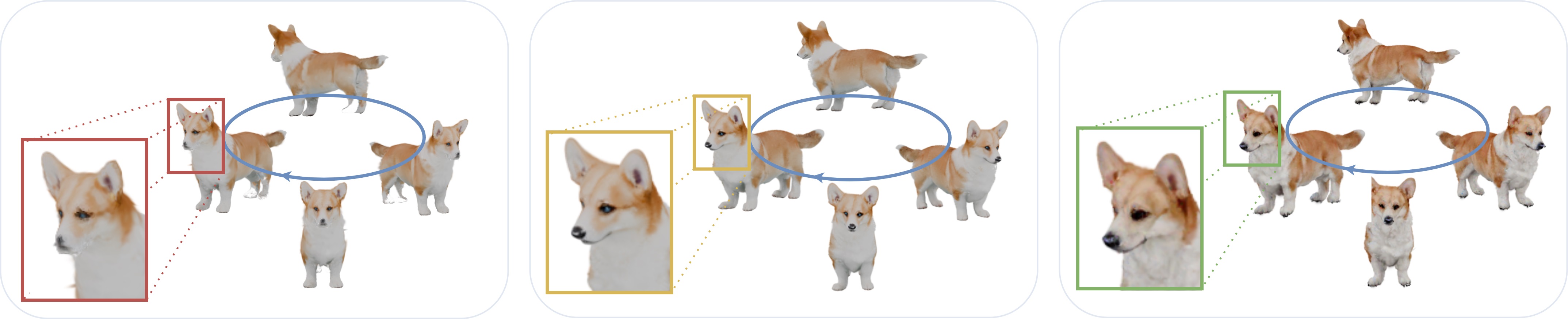}
\setlength{\tabcolsep}{4pt}
{\fontsize{6.8pt}{11pt}\selectfont
\begin{tabular}{ccc}
Stage 1  &  \hspace{3cm} Stage 2  \hspace{3cm} & Final \\
\end{tabular}
}
\caption{\our{} reconstructs synthetic data sets in two stages. Directly constraining Gaussian primitives to the mesh surface is insufficient and leads to visual artifacts. Stage 2 mitigates this limitation by enabling accurate refinement of geometry and appearance using unbounded Gaussians. Final to further eliminate the ,plastic'' effect and improve high-frequency details (e.g. fur), we apply DGE method, resulting in more natural and realistic reconstructions.}
\label{fig:stages}
\vspace{-0.6cm}
\end{figure}

\noindent\textbf{Model stages analysis} Fig.~\ref{fig:stages} and Tab.~\ref{tab:stage1dge} show the impact of stages on avatar quality.  We demonstrate that stages 1 and 2 are mainly used to the reconstruction of an asset generated using an image-to-3D model, here TripoSG.  Our results show that Stage 1 is insufficient to achieve faithful reconstruction.  Stage 2, which incorporates adaptive Gaussian density control, substantially improves reconstruction quality, It is nearly indistinguishable from the synthetic ground truth. Furthermore, the enhanced appearance modeling based on DGE consistently improves CLIP scores, indicating increased semantic alignment and improved visual quality.

\vspace{-0.3cm}
\section{Conclusion}
\vspace{-0.3cm}
In this work, we introduced \our{}, a novel framework for creating animatable 3D dog avatars from a single image. By combining the rendering power of 3D Gaussian Splatting with the mesh prior of the SMAL model, \our{} enables 3D reconstruction and controllable skeletal animation. Unlike existing image-to-3D methods, that typically produce non-parametric representations, our approach provides an articulated avatar, capable of motion transfer and editing. Our framework is highly flexible, supporting both direct pose manipulations and image-driven motion using FramePack and ActionMesh. Quantitative and qualitative comparisons demonstrate that \our{} achieves competitive visual quality and robust animation compared to current state-of-the-art baselines. 
\textbf{Limitations:}
Since our pipeline relies on an initial image-to-3D model to generate the pseudo-ground truth dataset, any topological errors are inherited. For instance, generative models sometimes produce incorrect geometry, such as a dog with three ears or extra limbs. While the Gaussians can represent these textures, it will not be a correct avatar corresponding to a real world dog. Additionaly, although ActionMesh provides a strong constraint, extreme camera rotations in the synthesized video can occasionally cause the tracker to lose alignment, leading to incorrect mesh in the rendered avatar.

\bibliographystyle{splncs04}

\clearpage

\renewcommand{\thesection}{\Alph{section}}
\setcounter{section}{0}
\section{Experiments}

\begin{wraptable}{r}{0.5\linewidth}
\vspace{-1.2cm}
\caption{Ablation on effect of specific losses on reconstruction quality. The table shows the average PSNR (dB) impact of removing individual loss terms.}
\label{tab:ablation_losses_psnr}
\begin{tabular}{l|cc|cc}
\hline
\multirow{2}{*}{Loss Term} & \multicolumn{2}{c|}{PSNR (dB)} & \multicolumn{2}{c}{$\Delta$ PSNR (dB)} \\
 & 15k & 40k & 15k & 40k \\ \hline
\textbf{Our (Full)} & \textbf{26.63} & \textbf{41.19} & - & - \\ \hline
w/o Dist & 26.60 & 41.17 & 0.03 & 0.02 \\
w/o Edge & 26.69 & 41.16 & -0.06 & 0.03 \\
w/o Lap. & 26.62 & 41.15 & 0.01 & 0.04 \\
w/o Offsets & 26.72 & 41.16 & -0.08 & 0.03 \\
w/o Pose & 26.74 & 41.18 & -0.11 & 0.01 \\
w/o Scale & 26.66 & 44.47 & -0.03 & -3.28 \\
w/o Opac & 26.63 & 42.23 & -0.00 & -1.04 \\
\hline
\end{tabular}
\end{wraptable}

\begin{figure}[!th]
    \begin{subfigure}[b]{\linewidth}
        \setlength{\tabcolsep}{2pt}
            \centering
            \begin{tabular}{lcc}
           w/o loss \hspace{1.0cm} & \hspace{1.0cm} Our (Full)
            \end{tabular}
            \includegraphics[width=0.7\linewidth]{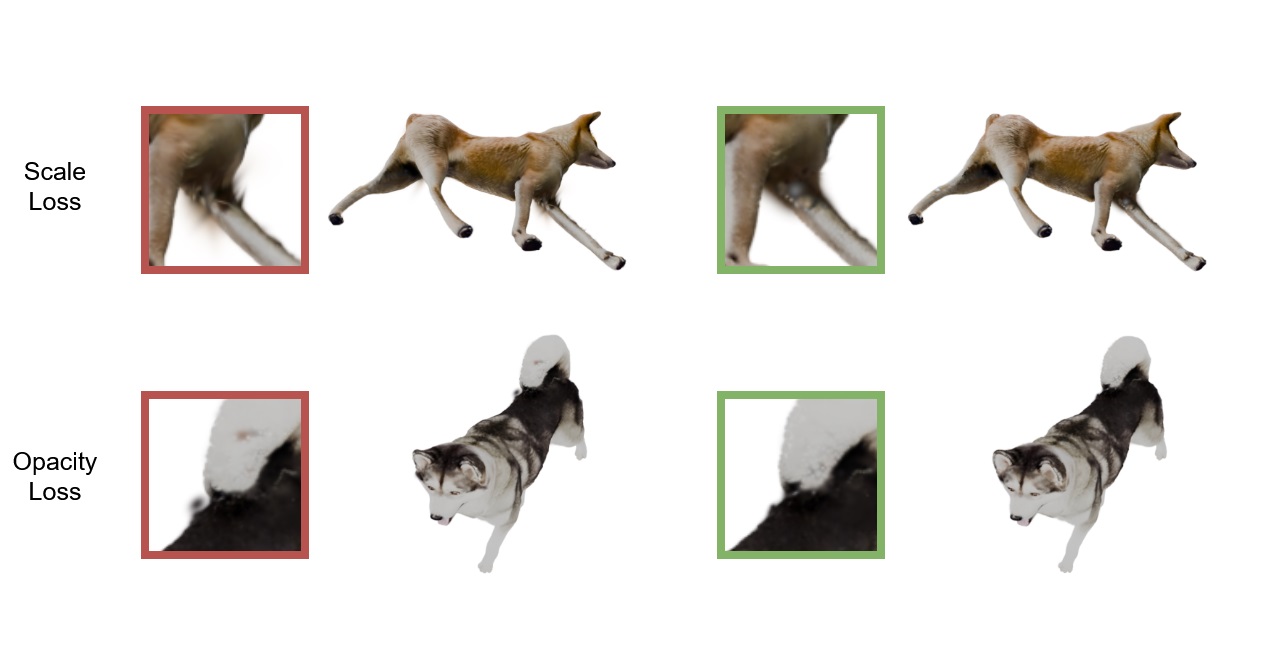}
        \caption{Impact of opacity $\mathcal{L}_{opac}$ and scale $\mathcal{L}_{scale}$ losses on visual quality during animations. In the first row spiky artifacts appear when the legs are moved if the opacity loss is disabled. The second row shows the noisy Gaussians visible during walking animation, visible near the dog's tail. The affected regions are highlighted.}
        \label{fig:losses_psnr}
    \end{subfigure}
    \begin{subfigure}[b]{\linewidth}
    \centering
        \setlength{\tabcolsep}{2pt}
            \begin{tabular}{lcc}
           w/o loss \hspace{1.0cm} & \hspace{1.0cm} Our (Full)
            \vspace{-1cm}
            \end{tabular}
            \includegraphics[width=0.7\linewidth]{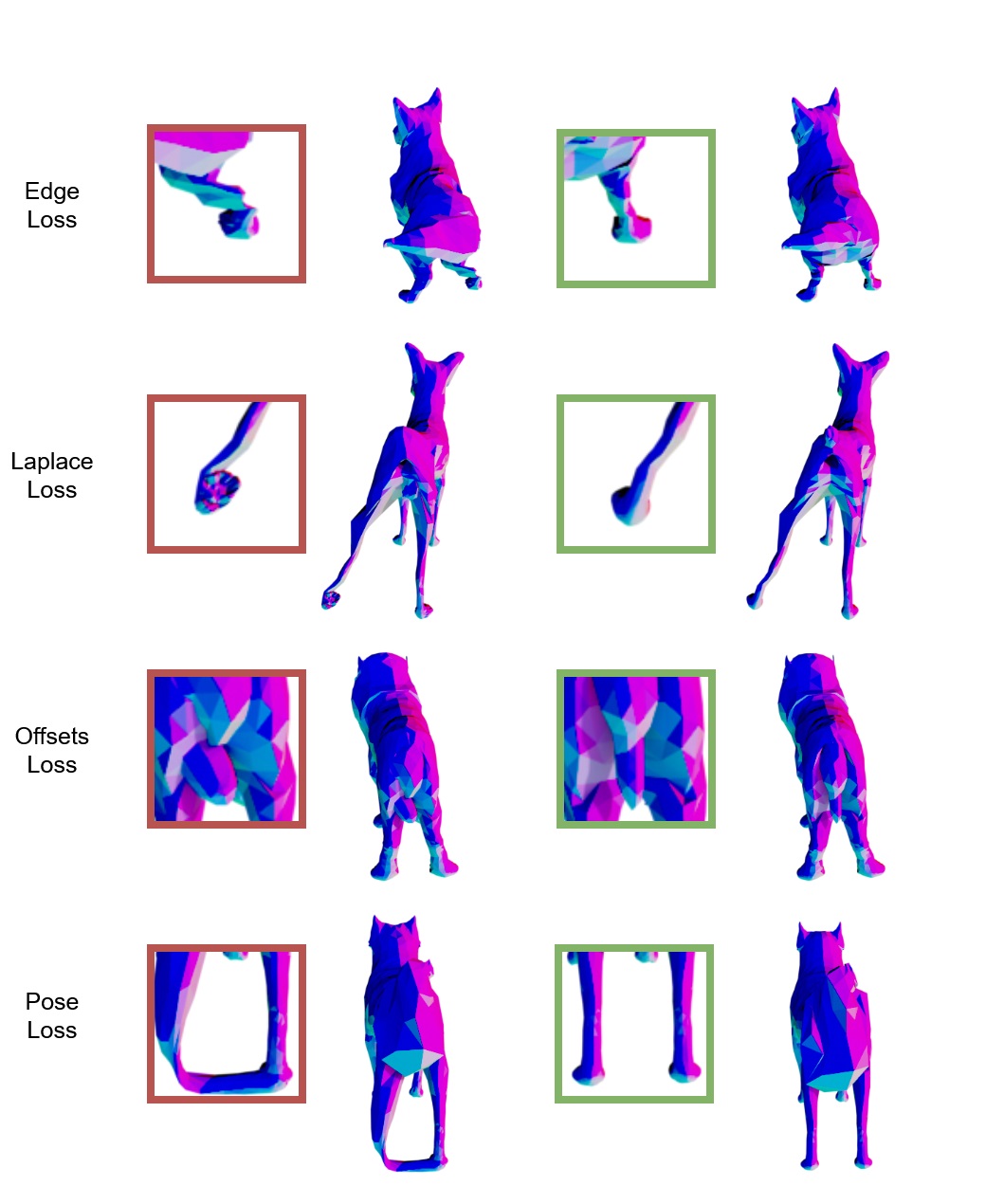}
        \caption{Impact of specific losses on the reconstructed SMAL mesh. The legs and tail are the most affected regions, and the artifacts include rough surface, twisting and mesh intersections.}
        \label{fig:losses_mesh}
    \end{subfigure}
    \caption{Visualization of ablation study on contribution of specific loss terms.}
\end{figure}

\textbf{Losses Ablation} We perform an additional ablation study on the impact of each loss term. We report the impact on reconstruction quality after Stage I (15k iterations), and Stage II (40k iterations) in Tab.~\ref{tab:ablation_losses_psnr}, measured by PSNR (dB) after disabling individual components. The inclusion of $\mathcal{L}_{opac}$ and $\mathcal{L}_{scale}$ provides a slight drop in PSNR. However, these terms are essential for the quality of animations, as illustrated in Fig.~\ref{fig:losses_psnr} removing them introduces noise and spiky artifacts respectively. Geometric regularizers $\mathcal{L}_{edge}, \mathcal{L}_{lap}, \mathcal{L}_{pose}, \mathcal{L}_{offsets}$ are critical for maintaining correct mesh geometry. As demonstrated in Fig.~\ref{fig:losses_mesh}, disabling these losses results in poor mesh topology. Their removal prevents the mesh from converging to an anatomically plausible geometry, which is crucial for mesh based skinning and animations. Moreover, their negative effect on reconstruction quality is negligible.

\noindent\textbf{Comparison with SOTA} In the main article we present comparisons with AnimalAvatar~\cite{sabathier2024animal}, GART~\cite{lei2024gart} and DogRecon~\cite{cho2025dogrecon}. The possibility of direct comparison with DogRecon is limited due to their code being unavailable. On the other hand, AnimalAvatar and GART try to reconstruct the avatar from a video, not a single image. This makes the comparison harder due to lack of exact SMAL parameters required to comparing exact camera view and pose. To conduct the experiments we chose random frames from the entire videos used to train GART and AnimalAvatar and applied image driven modifications or used the parameters from \textit{walking} sequence shared by GART. 

\section{Image Driven Modifications and Editability}

\begin{figure}[t]
    \centering
    \includegraphics[width=\linewidth, trim={0, 0, 0, 20}, clip]{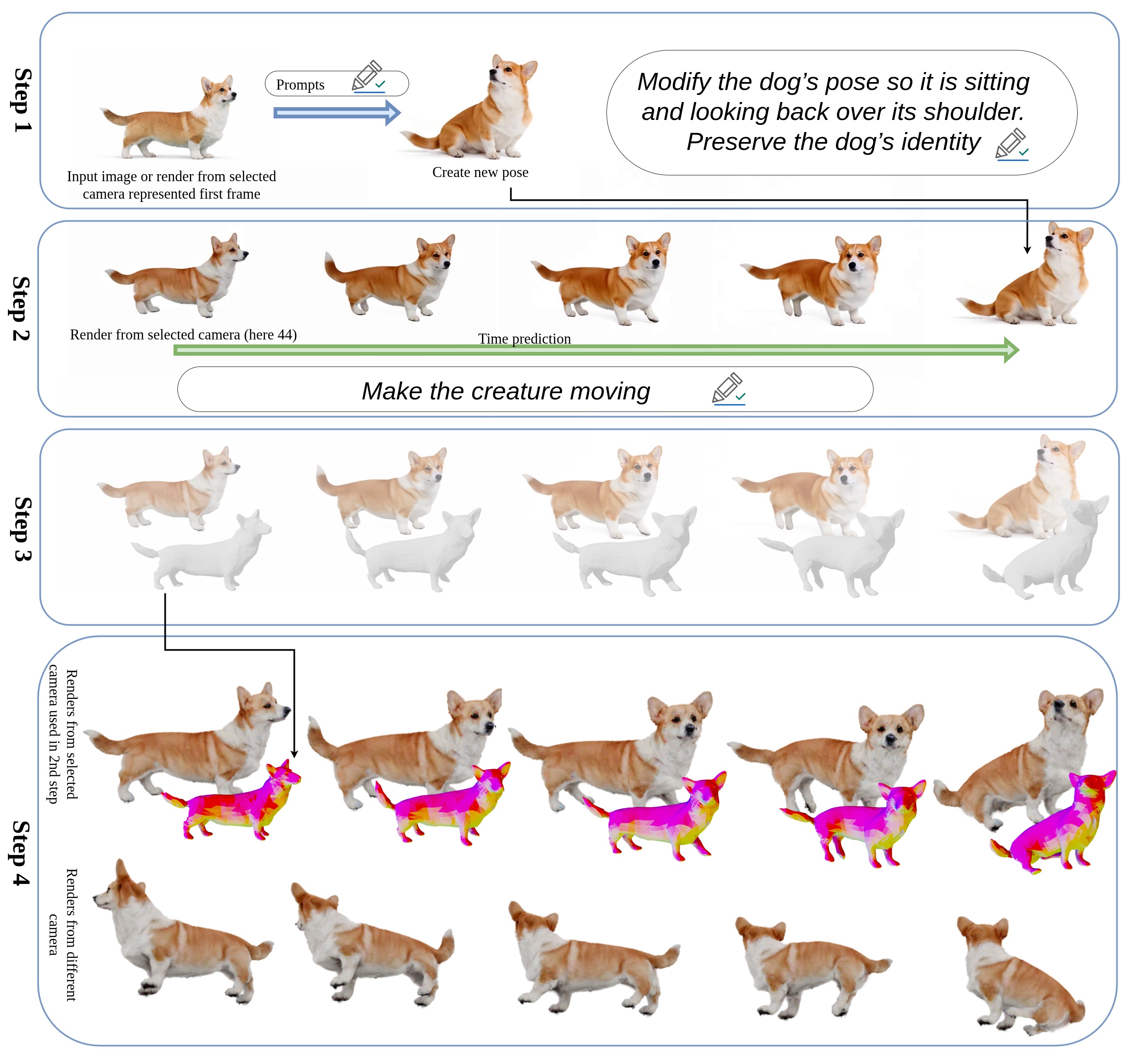}
    \caption{Recommended pipeline for Image-Driven Modifications. Our method enables pose and motion changes conditioned on text or image input. By operating on a consistent 3D representation, the \our{} supports rendering from multiple views with strong cross-view consistency and stable identity preservation of the dog. In this image, Step 1 takes the input image and generates the target image. In Step 2, the initial frame is captured from a selected camera view, which differs from the original input image.}
    \label{fig:textmodification}
\vspace{-0.75cm}
\end{figure}

This section describes the pipeline used to generate image driven modifications and semantic editing in our work. For the overview of image-based modifications see Fig.~\ref{fig:textmodification}. To create text- or image-guided animation, we recommend the following steps:
\\
\\
\noindent\textbf{Step 1: New pose generation} The first step is to construct the target image. In practice, the most convenient pipeline relies on image-to-image generation, where a source image is synthesized from a selected camera view or input image. The chosen camera should capture the dog as clearly and fully as possible. We recommend using image-to-image models.  We find both Qwen-Image-Edit-2511~\cite{wu2025qwenimagetechnicalreport} and GPT-4o~\cite{gpt4o} to be effective for this task. 

\begin{figure}[t]
    \centering
    \includegraphics[width=\linewidth]{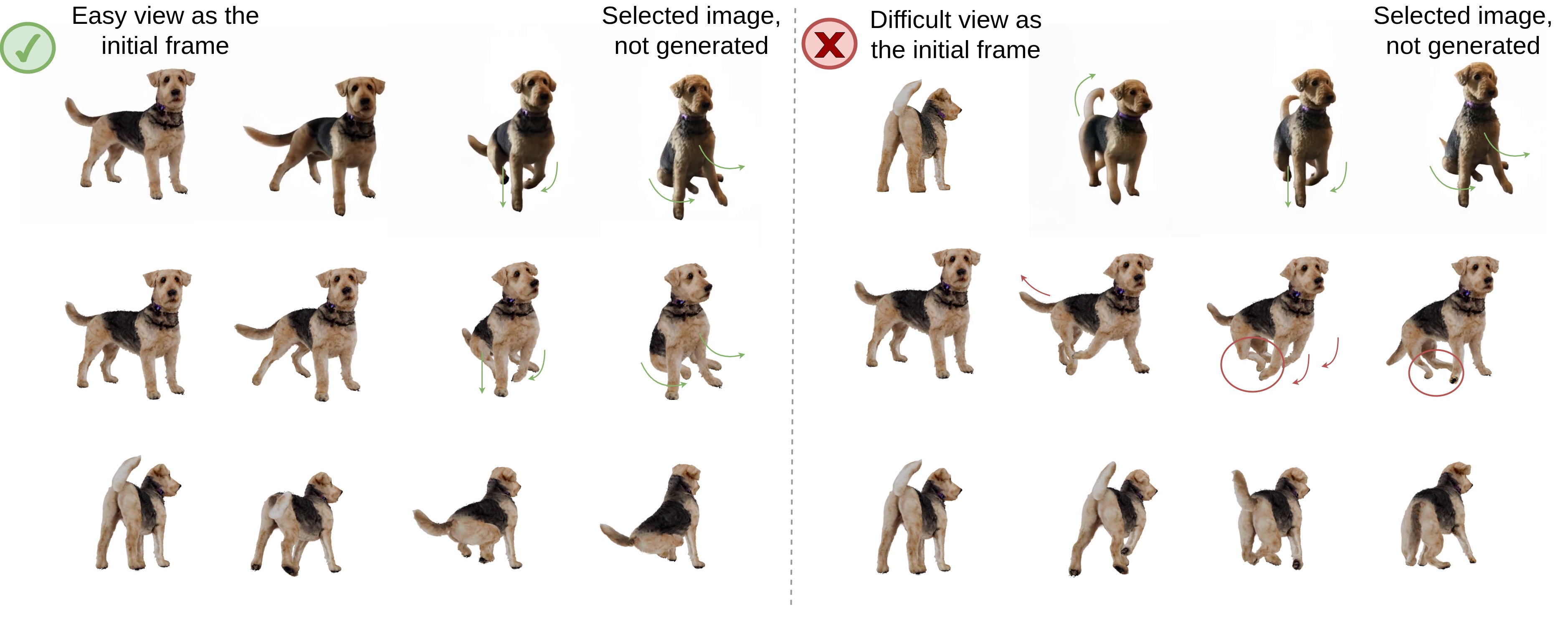}
    \caption{The impact of initial view on the creation of animations. Depending on the view ActionMesh might fail, leading to incorrect mesh geometry after deformation.}
    \label{fig:wrong_view}
\end{figure}

In this work, we also show an example where the target image was not generated but selected from a video so as to be able to compare it visually with other methods (Fig.~\ref{fig:coparision2}  in the main paper, Fig. \ref{fig:wrong_view} from appendix).

It is worth mentioning that selecting camera shots in which the dog is only partially visible should avoided. Such frame provide poor visual constraints and can lead to inconsistent or fitting poorly mesh, see Fig. \ref{fig:wrong_view}. An example of a failure scenario is discussed in the following paragraph.

\noindent\textbf{Step 2: Create a video} 
The next step is to generate a sequence of frames representing the motion. We assume that the first frame corresponds to the source image used in the first step. It may be the view from the selected camera or the input image itself. Based to the experiments, the best visual results are typically obtained when the source image used in the first step is used. However, this is not strictly required. The final frame of the sequence should correspond to the generated target image obtained in the first step. To generate a time-consistent video between two images simulation a video of the animal movement to the target pose FramePack~\cite{zhang2025frame} (HunyuanVideo~\cite{kong2024hunyuanvideo} transformer based version) is used. It is conditioned using the prompt ``the creature is moving.''. 

The generated target image as the final frame helps prevent visual artifacts. In contrast, generating a sequence using only the initial frame as guidance may lead to inconsistent motion across frames.

\begin{wrapfigure}{r}{0.5\textwidth}
 \centering
 \vspace{-1cm}
\setlength{\tabcolsep}{4pt}
{\fontsize{6.8pt}{11pt}\selectfont
\begin{tabular}{cc}
\hspace{0.8cm} Default \hspace{0.1cm} &  \hspace{0.7cm}  After box correction \\
\end{tabular}
  \includegraphics[width=0.9\linewidth, trim={0, 120, 120, 120}, clip]{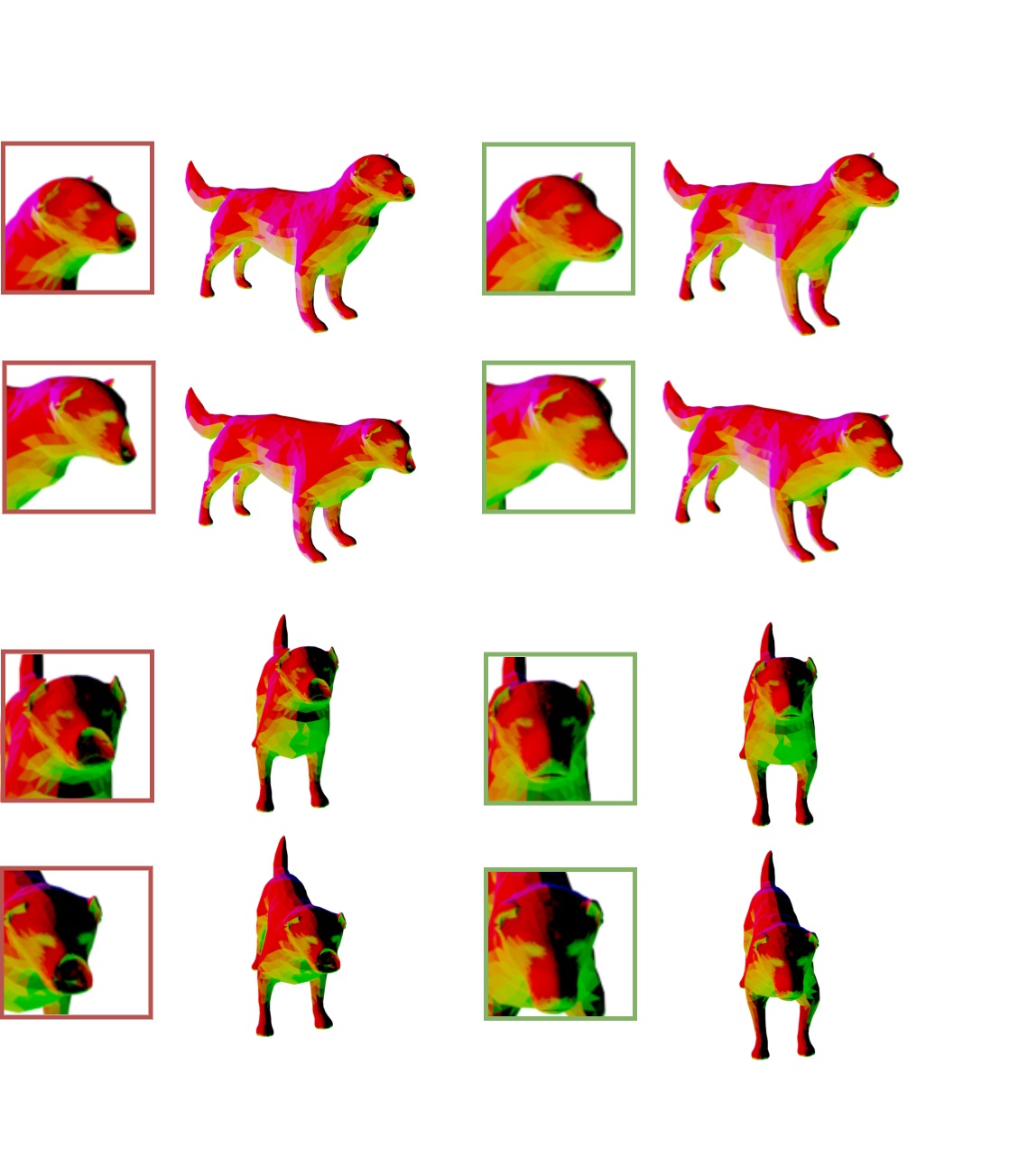}
}
\caption{Our method is compatible with ActionMesh. However, direct use can cause boundary artifacts. A bounding-box correction step is required.}
\label{fig:actionmes}
\vspace{-0.5cm}
\end{wrapfigure}
\noindent\textbf{Step 3: Fit Meshes} Our approach is compatible with ActionMesh \cite{sabathier2026actionmesh} using of SMAL mesh as a prior. In practice, animation is performed in a bounded box space, which may cause edge artifacts in some animation, see Fig. \ref{fig:actionmes} if the object is outside the box. To ensure complete compatibility with ActionMesh, 
the first frame (i.e. prior mesh) is scaled to a bounding box half, as it is in the original project.

\begin{figure}[t]
\centering
\setlength{\tabcolsep}{1pt}
{\fontsize{6.8pt}{11pt}\selectfont
\begin{NiceTabular}{cccccc}
  input & {sitting}  & \hspace{4mm} & 
    {walking}   & \hspace{4mm} & 
  {give a paw}  \\
\Block{3-1}{
\includegraphics[width=0.7cm]{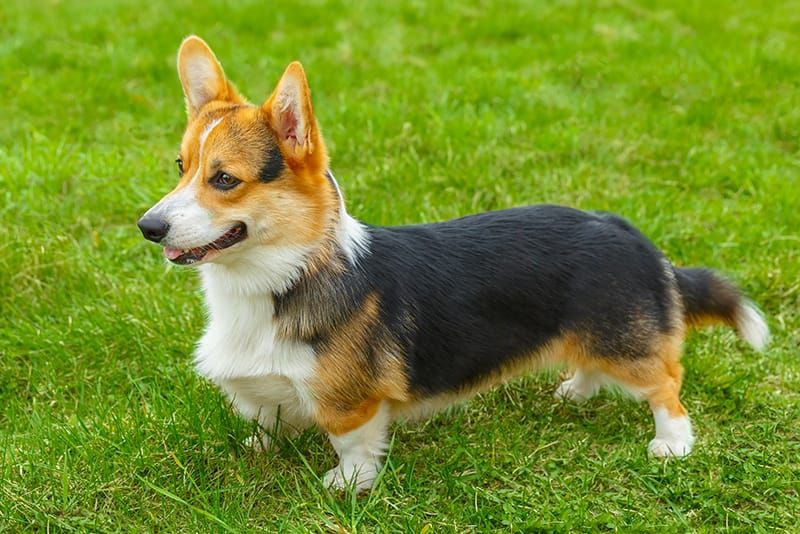}
} & 
\includegraphics[width=0.7cm]{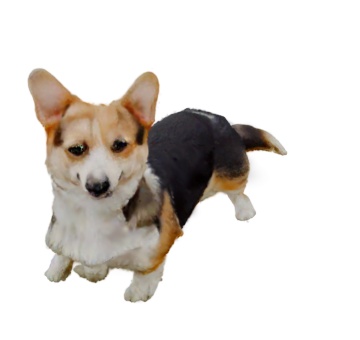} 
\includegraphics[width=0.7cm]{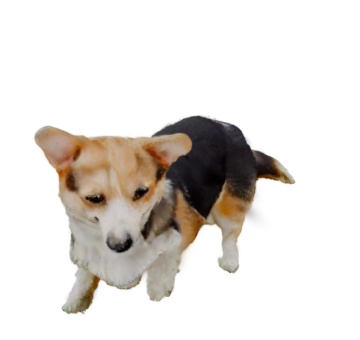} 
\includegraphics[width=0.7cm]{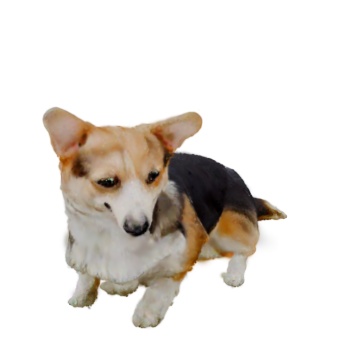} 
\includegraphics[width=0.7cm]{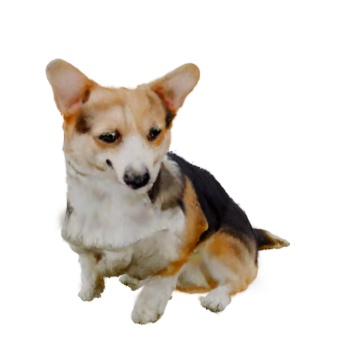}  & &
\includegraphics[width=0.7cm]{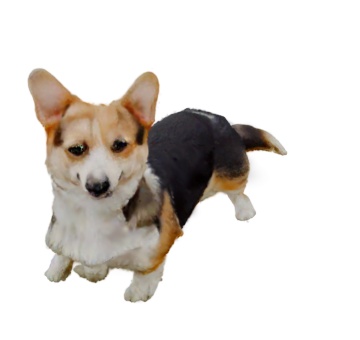} 
\includegraphics[width=0.7cm]{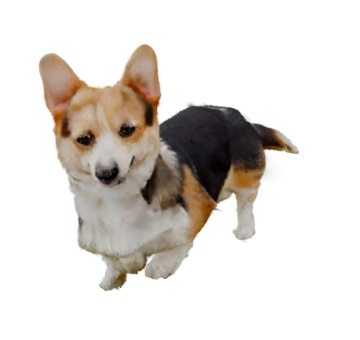} 
\includegraphics[width=0.7cm]{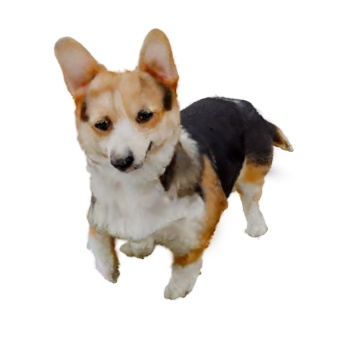} 
\includegraphics[width=0.7cm]{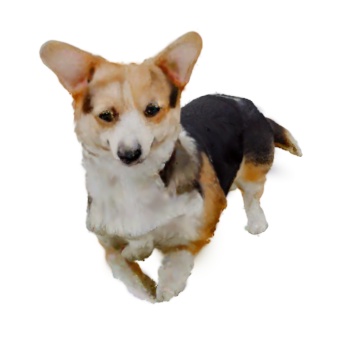}  & &
\includegraphics[width=0.7cm]{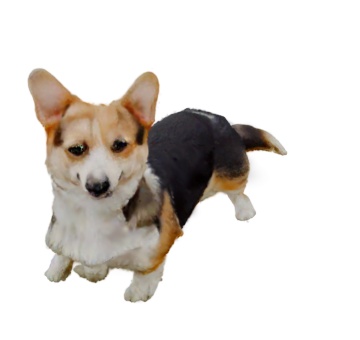} 
\includegraphics[width=0.7cm]{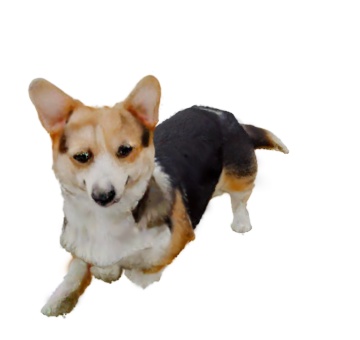} 
\includegraphics[width=0.7cm]{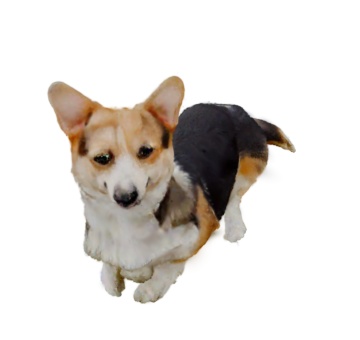} 
\includegraphics[width=0.7cm]{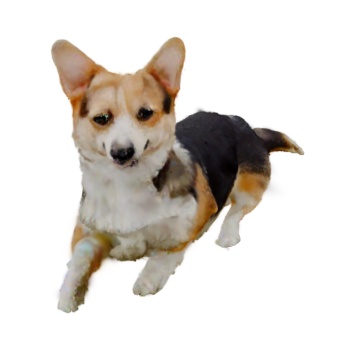}   \\
&
\includegraphics[width=0.7cm]{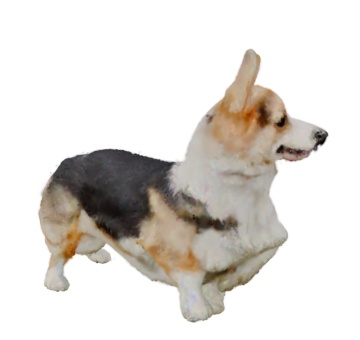} 
\includegraphics[width=0.7cm]{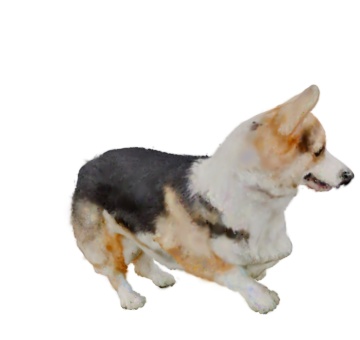} 
\includegraphics[width=0.7cm]{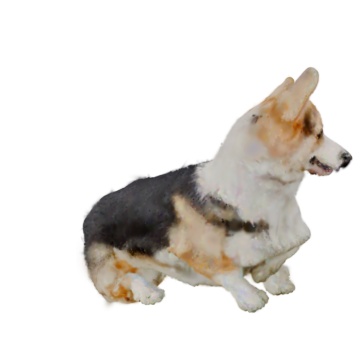} 
\includegraphics[width=0.7cm]{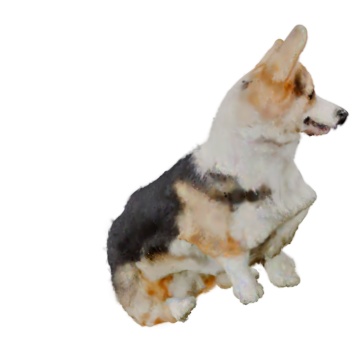}  & &
\includegraphics[width=0.7cm]{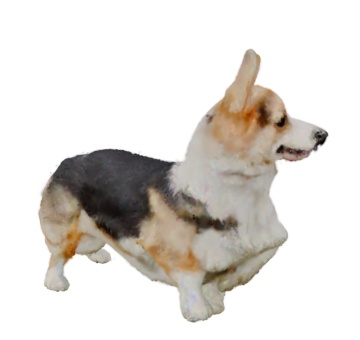} 
\includegraphics[width=0.7cm]{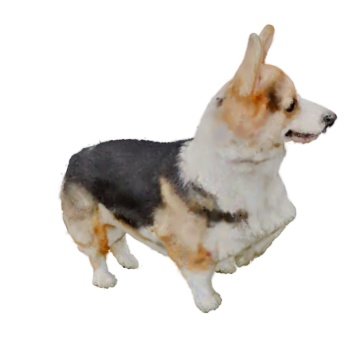} 
\includegraphics[width=0.7cm]{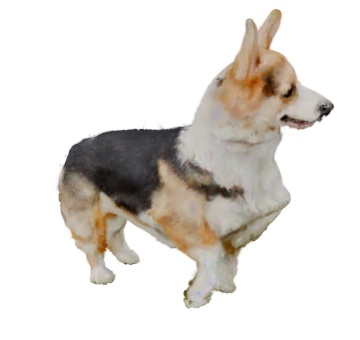} 
\includegraphics[width=0.7cm]{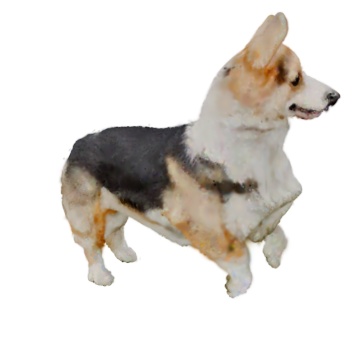}& &
\includegraphics[width=0.7cm]{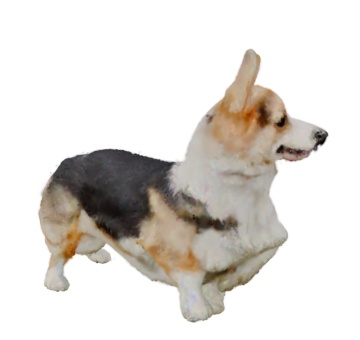} 
\includegraphics[width=0.7cm]{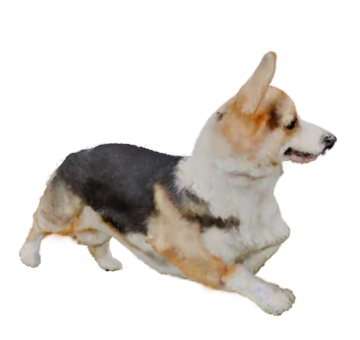} 
\includegraphics[width=0.7cm]{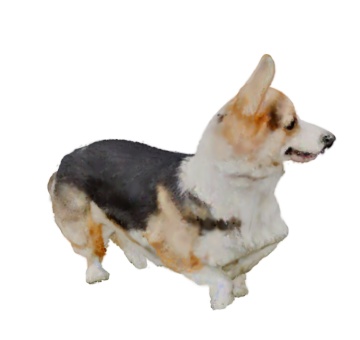} 
\includegraphics[width=0.7cm]{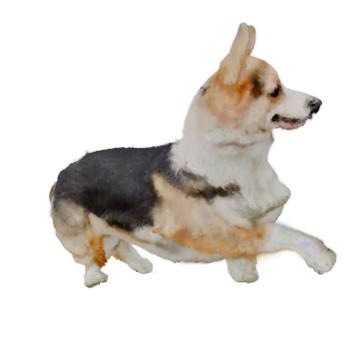}   \\
&
\includegraphics[width=0.7cm]{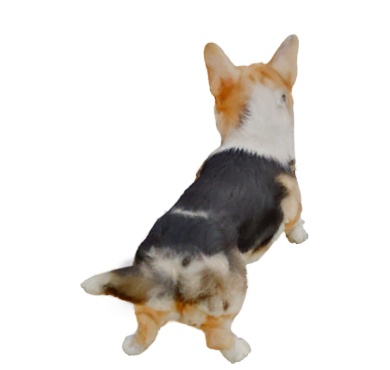} 
\includegraphics[width=0.7cm]{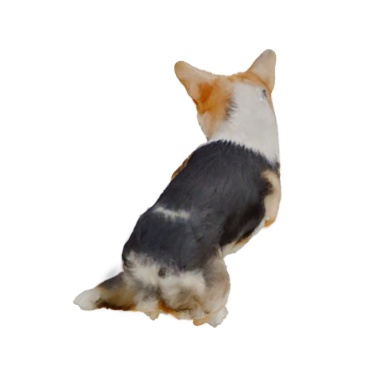} 
\includegraphics[width=0.7cm]{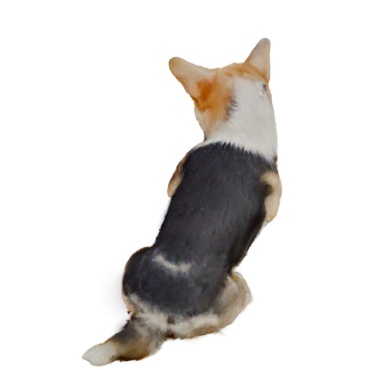} 
\includegraphics[width=0.7cm]{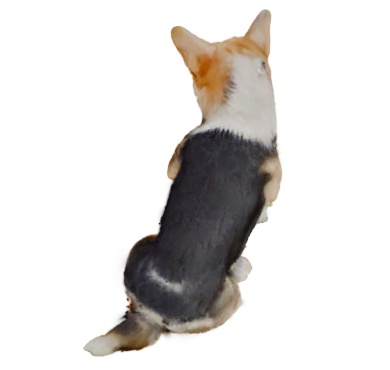}   & &
\includegraphics[width=0.7cm]{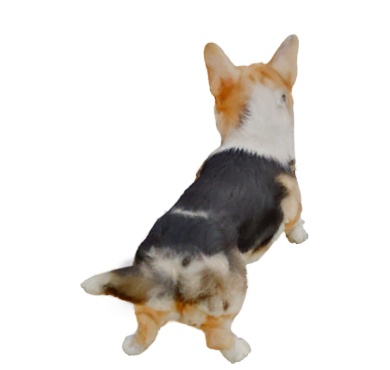} 
\includegraphics[width=0.7cm]{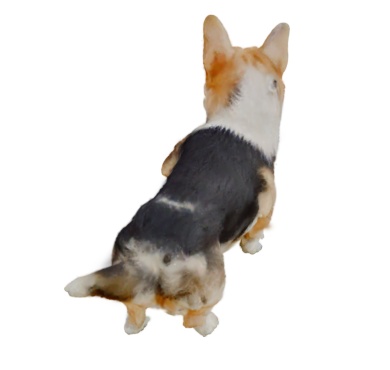} 
\includegraphics[width=0.7cm]{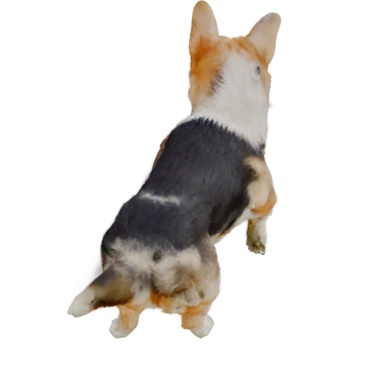} 
\includegraphics[width=0.7cm]{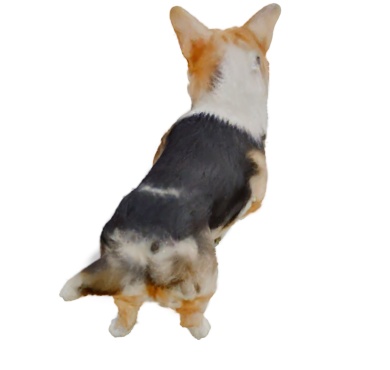}  & &
\includegraphics[width=0.7cm]{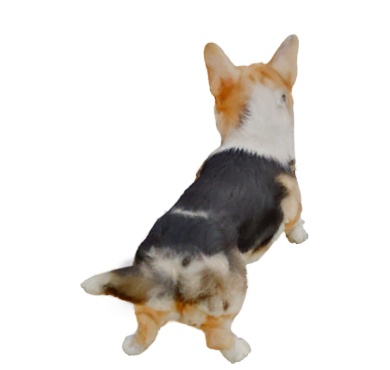} 
\includegraphics[width=0.7cm]{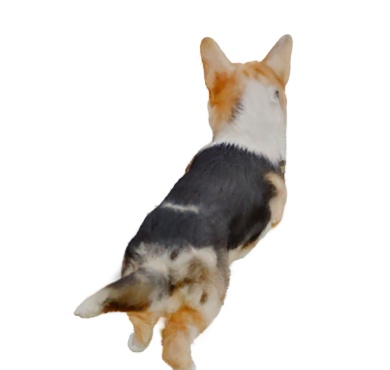} 
\includegraphics[width=0.7cm]{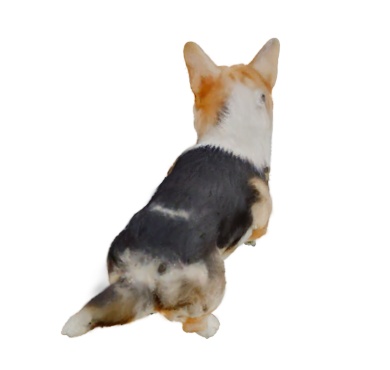} 
\includegraphics[width=0.7cm]{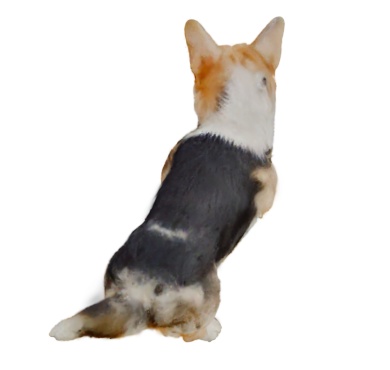}  
\\
&
\includegraphics[width=0.7cm]{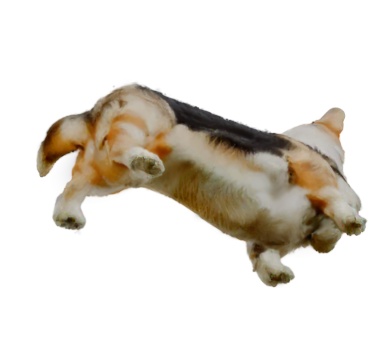} 
\includegraphics[width=0.7cm]{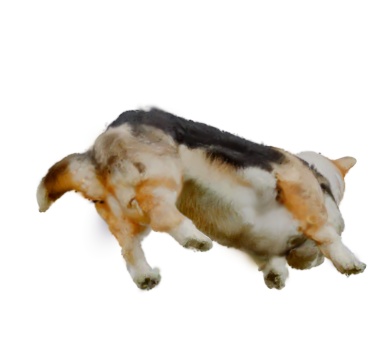} 
\includegraphics[width=0.7cm]{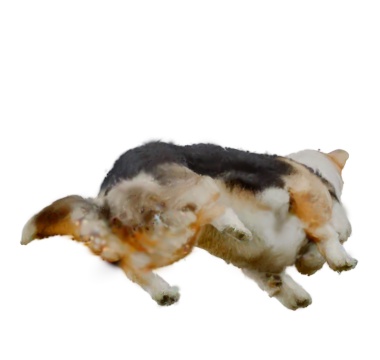} 
\includegraphics[width=0.7cm]{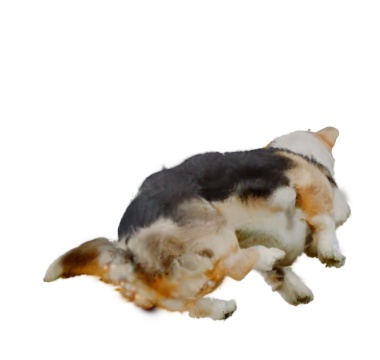}  & &
\includegraphics[width=0.7cm]{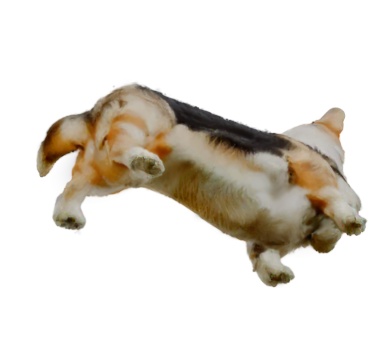} 
\includegraphics[width=0.7cm]{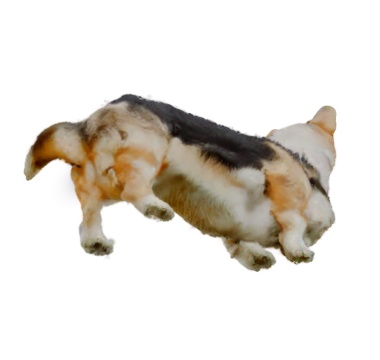} 
\includegraphics[width=0.7cm]{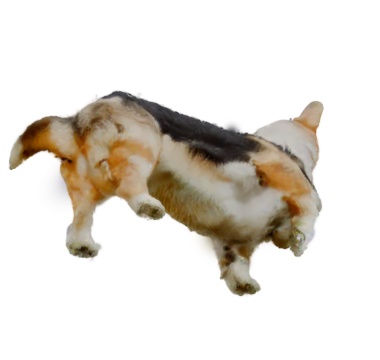} 
\includegraphics[width=0.7cm]{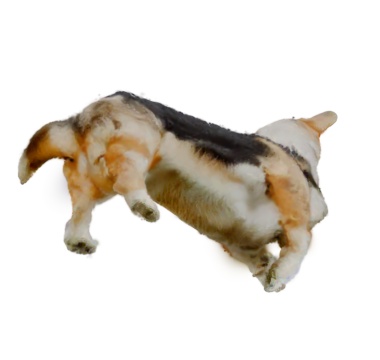}  & &
\includegraphics[width=0.7cm]{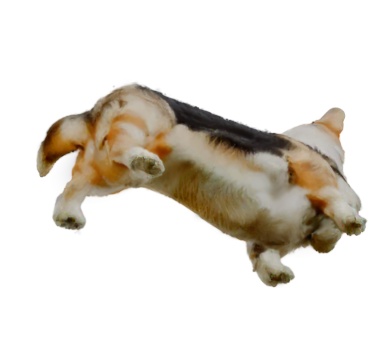} 
\includegraphics[width=0.7cm]{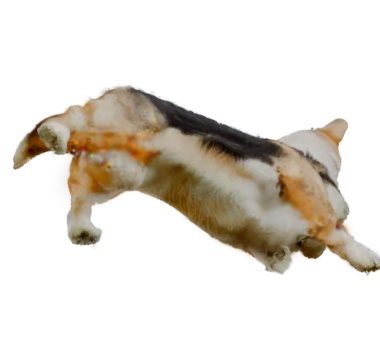} 
\includegraphics[width=0.7cm]{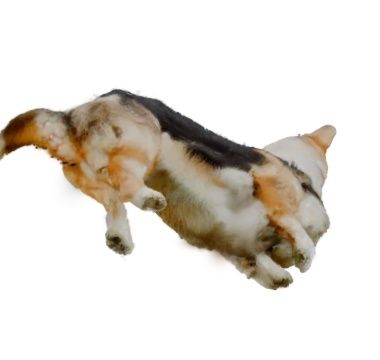} 
\includegraphics[width=0.7cm]{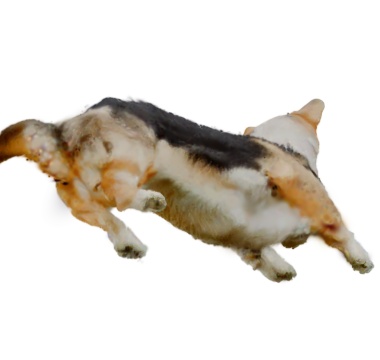}   \\

\end{NiceTabular}}
\caption{Additional avatar animations for arbitrary movements using Image Driven Modifications. The motion is created by generating a target frame using text-to-image generative model. It is used to interpolate intermediate frames using FramePack, which allow us to find corresponding mesh poses using ActionMesh.}
\label{fig:animations2}
\end{figure}

\noindent\textbf{Step 4: Mesh Transfer}
The represented movement of the output mesh sequence can be directly applied into our framework, as the Gaussians are attached (but not bounded) to the mesh faces. Consequently, as the mesh deforms, the Gaussians follow the same motion. This coupling ensures that renderings from different camera views consistently depict the same dog. it is worth noting that this causes preserving its identity, which was an issue that often arises in video-based generation approaches. The SMAL prior further helps maintain coherent animal geometry and characteristic properties throughout the motion.

\begin{wrapfigure}{r}{0.5\textwidth}
    \centering
    \vspace{-0.8cm}
    \includegraphics[width=\linewidth]{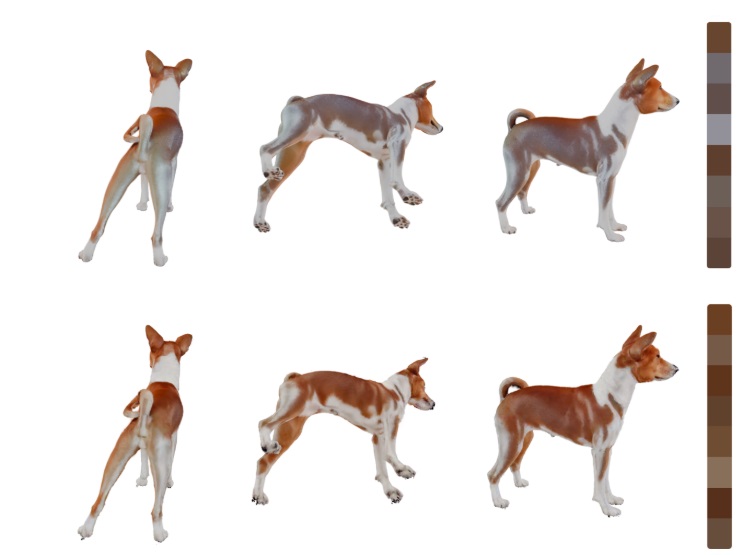}
    \caption{Comparison of the base reconstruction (first row) with the refinement after applying Direct Gaussian Editor (DGE) (second row). DGE significantly improves fur texture and photorealism.}
    \label{fig:dge}
    \vspace{-0.4cm}
\end{wrapfigure}

Fig.~\ref{fig:animations2} presents the continuation of the results from the Fig.~\ref{fig:animations} in the main article on the example of a corgi dog.

\noindent\textbf{Refinement via Direct Gaussian Editor}
The plastic-like look commonly occurs in generative models. We use the Direct Gaussian Editor (DGE)~\cite{chen2024dge} as a final post-processing step to mitigate this effect. Fig.~\ref{fig:dge} demonstrates this: the base reconstruction captures the overall structure but lacks fine-grained details, while the DGE-refined model exhibits significantly more realistic fur texture.



\end{document}